\pdfoutput=1
\documentclass{article}
\usepackage{iclr2027_conference,times}

\usepackage[utf8]{inputenc}
\usepackage[T1]{fontenc}
\usepackage{amsmath,amssymb,amsfonts}
\usepackage{amsthm}
\usepackage{mathtools}
\usepackage{graphicx}
\usepackage{booktabs}
\usepackage{float}
\usepackage{multirow}
\usepackage{xcolor}
\usepackage{colortbl}
\usepackage{bm}
\usepackage{algorithm}
\usepackage{algpseudocode}
\usepackage{microtype}
\usepackage{caption}
  \usepackage{enumitem}
  \setlist{itemsep=1pt, topsep=3pt, parsep=0pt, partopsep=0pt}
\usepackage{courier}
\usepackage{wrapfig}
\usepackage{tikz}
\usetikzlibrary{arrows.meta,positioning,calc,decorations.pathreplacing,fit,backgrounds}
\definecolor{mindon}{HTML}{2F6FAD}
\definecolor{mindoff}{HTML}{E7EBF1}
\definecolor{mindbad}{HTML}{C0392B}
\definecolor{mindink}{HTML}{3E4A59}
\definecolor{mindfaint}{HTML}{AEB8C4}

\tikzset{
  mcell/.style  ={draw=mindink, line width=0.25pt, inner sep=0pt,
                  minimum width=3.4mm, minimum height=3.4mm, fill=mindoff},
  mon/.style    ={mcell, fill=mindon},
  mbad/.style   ={mcell, fill=mindbad},
  mname/.style  ={font=\tiny, text=mindink, inner sep=0.6pt},
  msym/.style   ={font=\tiny, inner sep=0.6pt},
  mtitle/.style ={font=\scriptsize\bfseries, inner sep=0.6pt},
  mnote/.style  ={font=\tiny, text=mindink, inner sep=0.6pt},
  mred/.style   ={font=\tiny, text=mindbad, inner sep=0.6pt},
  marrow/.style ={-{Stealth[length=1.5mm,width=1.1mm]}, line width=0.55pt, draw=mindink},
  mthin/.style  ={line width=0.3pt, draw=mindink},
  mrule/.style  ={line width=0.3pt, draw=mindfaint},
  mpanel/.style ={draw=mindfaint, line width=0.3pt, rounded corners=0.8mm},
}
\usepackage{hyperref}
\usepackage{url}

\hypersetup{
  colorlinks=true,
  linkcolor=blue!70!black,
  citecolor=green!50!black,
  urlcolor=blue!60!black,
}

\title{Detecting and Explaining Robot Policy \\ Failures from Sparse Internal Features}

\author{Riad Ahmed, Sujosh Nag, Moniruzzaman Akash \& Momotaz Begum \\
CARL Lab, Department of Computer Science \\
University of New Hampshire, Durham, USA \\
\small\texttt{\{riad.ahmed, sujosh.nag, moniruzzaman.akash, momotaz.begum\}@unh.edu}}

\iclrfinalcopy

\begin{document}
\vspace*{-32pt}
\maketitle

\begin{abstract}
Visuomotor policies can handle many robot manipulation tasks, but understanding their failures
remains difficult.
Unfamiliar conditions and small action errors can lead to failure, yet watching the robot alone
tells us little about what went wrong inside the policy.
We introduce MIND (Mechanistic Interpretability for Neural Diagnosis), a method that looks inside a
trained policy to detect failures during execution and identify the internal concepts involved.
MIND uses a sparse autoencoder to break down the policy's internal activations into features that
can be studied individually.
We give meaning to these features by linking their activity to the robot's position, movement, and
surroundings, and by steering their activations to examine how they affect the robot's actions.
From successful and failed executions, we learn how the features behave on the way to a failure.
During execution, MIND flags deviations from these patterns and uses the meaningful features
contributing most to the change to describe the problem.
Experiments in simulation and on a physical robot show that changes in these internal features help
detect and explain failures.
By using the same features for detection and diagnosis, MIND connects failure signals to
understandable changes inside the policy.
\end{abstract}

\section{Introduction}
\label{sec:intro}
Mechanistic interpretability (MI) -- the study of a trained network to understand its
decision-making process through the features it represents instead of its individual neurons
\citep{elhage2022superposition, bricken2023monosemanticity, templeton2024scaling} -- has recently
gained popularity in robot learning to understand the internal operations of
vision-language-action (VLA) policies \citep{brohan2023rt2, kim2024openvla, black2024pi0,
pi05_2025}.
Some of these works use MI to `read' a policy's internal representation
\citep{lu2025probing, molinari2025emergent, drvla2025, jin2026event, grant2026features}, while
others use it to steer the policy in a closed-loop manner \citep{haon2025mechanistic,
buurmeijer2026observing}.
The former train linear probes \citep{alain2016probes} on the hidden layers of a VLA
\citep{lu2025probing, molinari2025emergent}, or decompose them into sparse-autoencoder features
whose meaning is then checked by clamping a feature while the robot runs \citep{drvla2025,
jin2026event, grant2026features}, whereas the latter use feed-forward directions or linear
observers to control what the robot does.
This paper belongs to the first group of work -- MIND clamps a feature in closed loop only to
verify that the policy acts through it \citep{templeton2024scaling}, never to control the robot.
The primary motivation to `read' a VLA policy's internal representation is to understand how and
why VLA policies -- which show remarkable performance on standard manipulation benchmarks such as
LIBERO \citep{liu2023libero} -- fail sharply under rather simple perturbations, such as a change in
the camera viewpoint or in the robot's initial state \citep{fei2025liberoplus}, and even within
their training distribution, where small action errors compound over a long horizon
\citep{ross2011dagger, ahmed2026tail}.
When such a policy fails, watching the robot shows what went wrong -- e.g., the gripper closing
short of the bowl -- but not why, because the process by which the policy turns an observation into
an action, hereafter referred to as the \textbf{Thought Process of the Policy} or \textbf{TPP},
stays hidden inside the network.

The major enabler of contemporary MI research is the sparse autoencoder (SAE)
\citep{bricken2023monosemanticity, cunningham2023sparse}.
An SAE reconstructs each activation vector as a sparse sum of learned \textsl{basis feature}
vectors \citep{elhage2022superposition}.
In language models many of these \textsl{basis features} respond to a single concept far more
selectively than individual neurons \citep{bricken2023monosemanticity, cunningham2023sparse,
gao2024scaling}, and amplifying or ablating one of them changes the model's output in a targeted
manner \citep{templeton2024scaling, marks2024sparse} -- much like adding a steering vector to the
activations \citep{turner2023steering}.
\citet{drvla2025} report a similar finding for VLAs, showing that SAEs decompose the internal
representations of $\pi_{0.5}$ and OpenVLA into largely interpretable and steerable
\textsl{basis features}.
This opens up an exciting new research direction where MI can be leveraged to look inside VLAs and
identify the changes in the TPP that accompany a task failure.
However, \citet{drvla2025} also conclude that most of these \textsl{basis features} are memorized
-- they fire on a narrow set of episodes or scenes instead of across many -- with at most one in
ten classed as general, and fewer than one in two hundred on OpenVLA.
Their SAE is fitted on the activations of all tokens of a frame averaged into one vector, and each
\textsl{basis feature} is named by a person who inspects the episodes on which it fires, so a name
is never tested on episodes outside that inspection.
In this paper, we argue that this conclusion depends on how the MI pipeline is constructed -- an
SAE fitted on every token separately, with each name selected by a fixed rule and kept only if it
still holds on held-out episodes, extracts many general
\textsl{basis features} from VLA policies. On the other hand, failure itself has been studied widely, and run-time detectors flag a failing execution from the
actions, the uncertainty or the internal features of a policy \citep{sentinel, fiper,
xu2025faildetect, gu2025safe, seligmann2026vlafail} -- SAFE, for instance, shows that the
activations of a VLA carry the information that a task is failing \citep{gu2025safe}.
Most of these detectors, however, compress that information into a single score, so their alarm
informs that a failure has occurred without identifying what in the TPP went wrong.

We propose MIND -- Mechanistic Interpretability for Neural Diagnosis -- a framework to track and
interpret the TPP in order to detect and explain VLA policy failures.
MIND 1) trains an SAE on the activations of the policy at every token during task execution,
2) gives each \textsl{basis feature} a name from a set of predefined concepts related to
manipulation tasks, keeps the name only if it still holds on held-out episodes, and, for a
decision concept, checks by clamping the \textsl{basis feature} in closed loop that the
policy acts through it, and 3) monitors these \textsl{basis features} during task execution,
raises an alarm when their activity departs from that of successful executions, and reports the
named \textsl{basis features} that contributed most to the alarm.
MIND therefore can explain a failure from the policy's own perspective -- for example, when the
gripper closes short of the bowl, an observer cannot tell whether the policy misjudged where the
bowl is or decided to close too early, but if the \textsl{basis feature} for the upcoming
\texttt{gripper-closing} decision is active while the \textsl{basis feature} for the
\texttt{gripper-near-bowl} state remains inactive, MIND reports the failure as an `early closure
decision' instead of a `poor estimation of the bowl position'.
Extensive evaluation of MIND in simulation with OpenVLA \citep{kim2024openvla}, $\pi_{0.5}$
\citep{pi05_2025} and LingBot-VLA \citep{wu2026lingbot}, and on a physical robot running
$\pi_{0.5}$ demonstrates its performance in detecting and explaining VLA policy failures.
In OpenVLA, clamping a single named \textsl{basis feature} raises the share of steps that move the
named way from $0.00$--$0.25$ to $0.92$--$1.00$ for twelve of fourteen motion concepts, and in all three policies the named \textsl{basis features} detect failures as well as
the dense activation -- an AUROC of $0.991$ against $0.988$ on $\pi_{0.5}$, with half the false
alarms.
Our contributions are:
\begin{itemize}
\item We extract sparse basis features of a VLA whose names are selected on one set of runs and remain
  precise on unseen initial states and on an unseen task, where the neurons, residual coordinates
  and random directions of the same policy provide none (Sections~\ref{sec:method}
  and~\ref{sec:results}).
\item We verify in closed loop that the policy acts through these basis features, since clamping one
  basis feature steers the arm while random ones of equal magnitude do not
  (Section~\ref{sec:results}).
\item We build a run-time monitor that raises an alarm from the same basis features and identifies the
  internal quantities responsible for it, evaluated on three policies in simulation and on a
  physical robot (Sections~\ref{sec:setup} and~\ref{sec:results}).
\end{itemize}
Unlike a number of recent works on policy safety
\citep{zhang2026probeact, sinha2024realtime, liu2023reflect, ahmed2026tail}, MIND does not offer a
way to recover from failure.
Rather, the goal of MIND is to understand what the policy represented when it failed -- as opposed
to just informing that a failure has occurred -- which will in turn facilitate designing a recovery
mechanism. 

\section{Method}
\label{sec:method}
\label{sec:method:system}
\begin{figure}[t]
\centering \includegraphics[width=\textwidth]{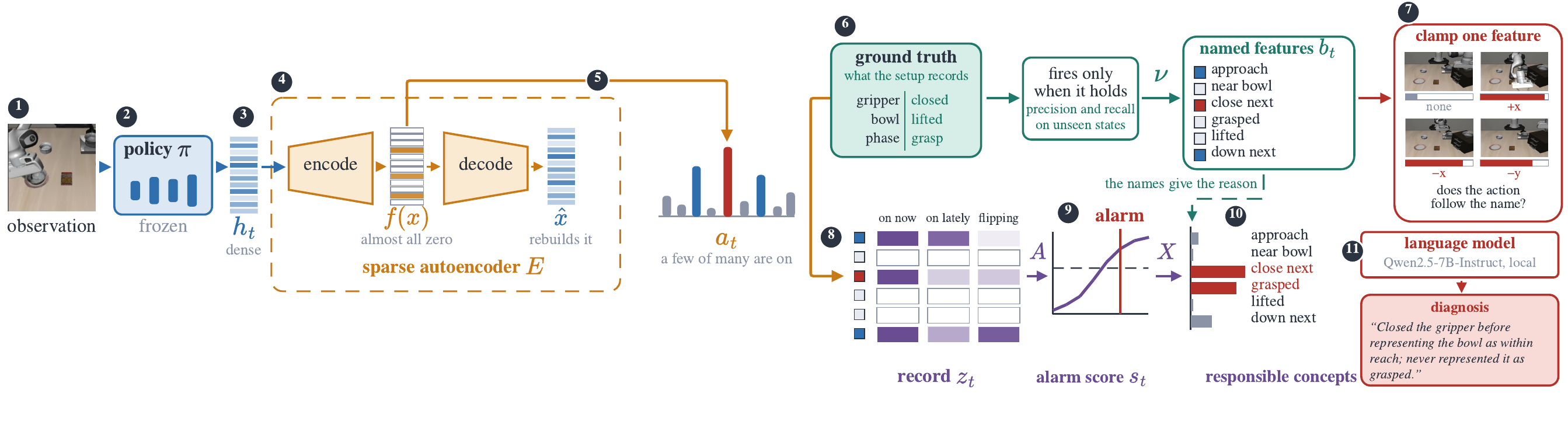} \caption{MIND end to end. The policy runs untouched (1--2), and a sparse autoencoder turns one of its
activations into a sparse code (3--5). Top: a basis feature is named only if a
recorded fact holds almost every time it fires (6), and a clamp tests whether the policy acts through it (7). Bottom: the basis features
form a record at every step (8), a linear alarm scores it (9), the score splits over the concepts that
moved it (10), and a language model puts them into words (11). Frames are recorded rollouts; traces
follow the toy example of this section.}
\label{fig:system}
\end{figure}

MIND decomposes the TPP of a VLA policy into basis features with human-readable names, and it detects and
explains a failing attempt through those same basis features.
The basis features come from the \textbf{activation} $h_t$, the vectors that one layer of the policy
returns at step $t$.
The method rests on two conditionals:
(i) if each activation decomposes into numbers of which only a few are non-zero, each number follows
one quantity of the TPP;
(ii) if each number also has a human-readable name that holds on unseen attempts at run time, the
TPP at every step reads as the list of named basis features that are on.
Failure detection is a product of the two: a score computed from the same numbers separates a
failing attempt from a successful one, and tracing that score back to the numbers and their names
reports the failure from the policy's own perspective.
Figure~\ref{fig:system} draws the pipeline on the running task, in which the robot picks up a black
bowl and places it on a plate.
After the preliminaries of Section~\ref{sec:method:prelim}, we construct the autoencoder and the
naming rule of MIND that satisfy (i) and (ii) in Sections~\ref{sec:method:sae}
and~\ref{sec:method:name}, and build the failure detector and its explanation on them in
Sections~\ref{sec:method:monitor} and~\ref{sec:method:cause}.

\subsection{Preliminaries}
\label{sec:method:prelim}

In a VLA policy $\pi$, an episode is one attempt at a task, from the initial state of the scene to
the goal or the step limit.
At each step $t$ the policy receives frame $t$, the current camera image, and a fixed instruction in
words.
Each layer of the policy writes one vector of width $d$ per position, where a position, or token, is
one patch of the image, one word of the instruction or one part of the coming action.
The activation $h_t$ that layer $\ell$ returns is dense, with almost every coordinate non-zero at
almost every step.
No single coordinate of $h_t$ has a meaning that a human can name, such as the gripper grasping the
black bowl.
The reason is superposition \citep{elhage2022superposition}.
Because a policy represents far more quantities than a layer has coordinates, each coordinate mixes
parts of many quantities, and one quantity such as the gripper action spreads over several
coordinates.

\textbf{Sparse autoencoder:} A sparse autoencoder separates these quantities into separate directions
\citep{bricken2023monosemanticity, cunningham2023sparse}.
It learns $m$ \textsl{basis features}, each with a
direction in $\mathbb{R}^{d}$.
Because $m$ is larger than $d$, each quantity that the policy follows to compute an action can
receive a separate basis feature.
Together these directions form the \textbf{dictionary} of sparse coding \citep{olshausen1997sparse}.
The direction of basis feature $i$ is a unit vector $u_i \in \mathbb{R}^{d}$ that identifies the basis feature,
with a length $\gamma_i > 0$ that scales how strongly it is on.
The \emph{encoder} of the sparse autoencoder $E$ maps one vector $x \in \mathbb{R}^{d}$ at a time to
its \textbf{code} $f(x)$, with one number per basis feature, and the \emph{decoder} rebuilds from it a
reconstruction $\hat{x}$,
\begin{equation}
f(x) = \mathrm{ReLU}\bigl(W_{\mathrm{enc}}\,x + b_{\mathrm{enc}}\bigr) \in \mathbb{R}^{m},
\qquad
\alpha_i(x) = \gamma_i\, f_i(x),
\qquad
\hat{x} = b_{\mathrm{dec}} + \sum_{i=1}^{m} \alpha_i(x)\, u_i .
\label{eq:sae}
\end{equation}
In Equation~\ref{eq:sae} the ReLU lets a number of the code be exactly zero, and the decoder adds each
direction $u_i$ to its bias $b_{\mathrm{dec}}$, weighted by the \textbf{basis-feature activation}
$\alpha_i(x)$.
The code is sparse: almost every number in it is zero, and the few that are not belong to the
basis features that are \emph{on} at $x$.
The encoder and the decoder are fitted together, without labels, on activation vectors recorded over
a large set of inputs \citep{bricken2023monosemanticity, templeton2024scaling}, by minimizing
\begin{equation}
\mathcal{L}(x) = \underbrace{\lVert x - \hat{x} \rVert_2^2}_{\text{reconstruction}}
\; + \; \lambda \underbrace{\sum_{i=1}^{m} \alpha_i(x)}_{\text{sparsity}} ,
\label{eq:saeloss}
\end{equation}
where a larger $\lambda$ leaves fewer basis features on at each vector.
Both terms stay small when each basis feature follows one quantity of the policy, because such a quantity
is present at a few vectors and its one direction rebuilds it wherever it appears.

\textbf{Naming:} A basis feature found this way is a direction, and naming gives it a meaning.
The name of a basis feature is a \textbf{concept}, a human-readable fact such as the gripper closing on the
black bowl.
In prior work a person reads the inputs at which the basis feature is most strongly on, its
\emph{top-activating examples}, and writes down what they have in common as the name
\citep{bricken2023monosemanticity, templeton2024scaling}.
Such a name says that the basis feature is on when the concept is true.
Whether the model also uses the basis feature is tested by \textbf{steering}, which fixes the basis feature
activation at a large value while the model runs.
The name is confirmed when the output then changes the way the name predicts
\citep{templeton2024scaling}.

\subsection{The sparse autoencoder of MIND}
\label{sec:method:sae}

MIND builds the code $f_t$ of every step with a sparse autoencoder $E$ fitted on the activations of
the policy itself.
$E$ encodes the vector at each position of $h_t$ that it reads, and all positions share one
dictionary.
Over the positions $\mathcal{P}$ of step $t$ these codes form
\begin{equation}
f_t \;=\; \bigl[\, f\bigl(h_t^{(p)}\bigr) \,\bigr]_{p \in \mathcal{P}}
\;\in\; \mathbb{R}^{m \times \lvert \mathcal{P} \rvert},
\label{eq:ft}
\end{equation}
where column $p$ is the code of $h_t^{(p)}$, the vector of $h_t$ at position $p$.
In Equation~\ref{eq:ft} the parts of the TPP stay apart, since the policy writes the coming action at
the action positions, keeps the instruction at the words and the scene at the image patches.
The position at which a basis feature is on thus becomes part of its meaning.
When the gripper is about to close on the black bowl, for example, the basis features that turn on at the
action positions follow that closure.
Because most vectors of a frame are image patches, $m$ is several times larger than $d$
(Appendix~\ref{app:sae}), which leaves basis features for the words and the action positions as well.

The layer $\ell$ that $E$ encodes lies near the middle of the stack, where the scene, such as the gripper above the
black bowl, is well represented.
We fit its dictionary on recorded episodes, successful or failed, which we call the \emph{fitting
episodes}, and the positions of their frames enter the fit,
\begin{equation}
\min_{E}\; \sum_{e \in \mathcal{D}} \,\sum_{t \in \mathcal{S}_e} \,\sum_{p \in \mathcal{P}}
\mathcal{L}\bigl(h_t^{(p)}\bigr),
\label{eq:fit}
\end{equation}
where $\mathcal{D}$ is the set of fitting episodes and $\mathcal{S}_e$ the frames of episode $e$ used
for the fit.
A policy with a separate action expert has a second stream of vectors of a different width, and MIND
fits one autoencoder per stream that it reads.
Neither fit refers to the concepts, the task or how an episode ended (Appendix~\ref{app:sae} gives
the choice of $\lambda$).
Fitted this way, many basis features of $f_t$ keep one meaning on episodes outside the fit and can be named.
Throughout the rest of the section we follow a toy example of this code, whose activation has one
position with two coordinates and whose dictionary has six directions.
Its code has six numbers, and we call it the \textbf{toy vector} $g_t$ to keep it apart from the code
$f_t$ of a real policy.
At a frame where the arm is still coming down toward the black bowl it takes the value
$g_t = (0.8,\, 0,\, 1.2,\, 0,\, 0,\, 0.5)^{\top}$, chosen for illustration, and its three zeros are
the sparsity that Equation~\ref{eq:saeloss} asks for.

\subsection{Naming of a basis feature}
\label{sec:method:name}

Before any activation vector is examined, we fix the set $\mathcal{C}$ of concepts.
It contains \textbf{state concepts} such as the bowl grasped and \textbf{decision concepts} such as the
gripper closing next.
Names are chosen from $\mathcal{C}$ on the \emph{naming frames}, the frames of the fitting episodes
that Equation~\ref{eq:fit} did not use, and tested on \emph{held-out episodes}, recorded episodes
outside the fitting episodes (Section~\ref{sec:setup} and Appendix~\ref{app:setup} give this split
and the source of the concepts).
Whether a basis feature can have a name depends on how strongly and how often it is on with a concept.
Basis feature $i$ \emph{fires} on $x$ when $\alpha_i(x)$ exceeds a firing level $\tau\,\alpha^{\max}_i$,
where $\alpha^{\max}_i$ is the largest value that the basis feature reached during the fit and
$\tau \in [0,1)$.
A small $\tau$ counts weak activations as firings, and a larger $\tau$ only the strong ones.
The weak firings of a basis feature spread over many kinds of frames, while its strong firings gather on a
few frames that look alike \citep{templeton2024scaling}.
We therefore try every $\tau$ in a small fixed set $\mathcal{T}$ and make the level part of the name.

Let $\mathcal{F}^{\tau}_i$ be the set of vectors on which basis feature $i$ fires at level $\tau$ and
$\mathcal{X}_c$ the set on which concept $c$ is true.
Precision and recall measure how well a pair $(c,\tau)$ of a concept and a level fits basis feature $i$,
\begin{equation}
P_i(c,\tau) = \frac{\lvert \mathcal{F}^{\tau}_i \cap \mathcal{X}_c \rvert}{\lvert \mathcal{F}^{\tau}_i \rvert},
\qquad
R_i(c,\tau) = \frac{\lvert \mathcal{F}^{\tau}_i \cap \mathcal{X}_c \rvert}{\lvert \mathcal{X}_c \rvert} .
\label{eq:precrec}
\end{equation}
In Equation~\ref{eq:precrec} precision $P_i$ is the share of the basis feature's firings on which the
concept is true, and recall $R_i$ the share of the concept's vectors on which the basis feature fires.
A pair is allowed for basis feature $i$ when its precision reaches a floor $p_{\min}$, its recall reaches a
floor $r_{\min}$ and the basis feature has fired at least $n_{\min}$ times at that level.
The first floor makes the concept true almost every time the basis feature fires, and the second keeps out
a concept that the basis feature meets by chance.
Among the allowed pairs $\mathcal{A}_i$ the name $\nu(i)$ is the one with the highest recall,
\begin{equation}
\begin{aligned}
\nu(i) &= \operatorname*{arg\,max}_{(c,\tau)\,\in\,\mathcal{A}_i} R_i(c,\tau), \\
\mathcal{A}_i &= \bigl\{ (c,\tau) \in \mathcal{C}\times\mathcal{T} :
P_i(c,\tau) \ge p_{\min},\; R_i(c,\tau) \ge r_{\min},\;
\lvert \mathcal{F}^{\tau}_i \rvert \ge n_{\min} \bigr\} .
\end{aligned}
\label{eq:name}
\end{equation}
A basis feature that Equation~\ref{eq:name} gives a name is \emph{named}, and only named basis features enter
the monitor of Section~\ref{sec:method:monitor}.
On the held-out episodes the name and its level stay fixed.
The name \emph{holds} when precision and recall clear the same floors there and the basis feature has fired
at least $n_{\min}$ times, spread over a fixed number of episodes with no single episode supplying
more than a fixed share (Appendix~\ref{app:name}).
A basis feature whose name holds is monosemantic, since it fires with one concept on episodes that took no
part in choosing its name.
Figure~\ref{fig:name} walks $g_3$ through this rule.

Under this rule the six basis features of the toy vector take the names that Figure~\ref{fig:course} draws
as its rows.
Four of them take state concepts: $g_1$ the arm approaching the bowl, $g_2$ the gripper near the bowl,
$g_4$ the bowl grasped and $g_5$ the bowl lifted.
The other two take decision concepts: $g_3$ the gripper closing next and $g_6$ the arm going down
next.
At the frame of Section~\ref{sec:method:sae} the basis features on are $g_1$, $g_3$ and $g_6$.
They say that the policy is on its way down and about to close the gripper while the gripper is not
yet near the bowl.
A name earned this way says only that the basis feature is on when its concept is true, and for a decision
concept we also test by steering whether the policy writes the action through that basis feature.
Here the clamp is a fixed multiple of $\alpha^{\max}_i$ at every step of a closed-loop rollout, and
for the toy vector the test asks whether clamping $g_6$ sends the arm down.
Appendix~\ref{app:setup} gives the rollouts, the clamped basis features and the controls, and
Appendix~\ref{app:name} the multiple.

\subsection{Failure detection}
\label{sec:method:monitor}

\begin{figure}[t]
\centering \includegraphics[width=\textwidth]{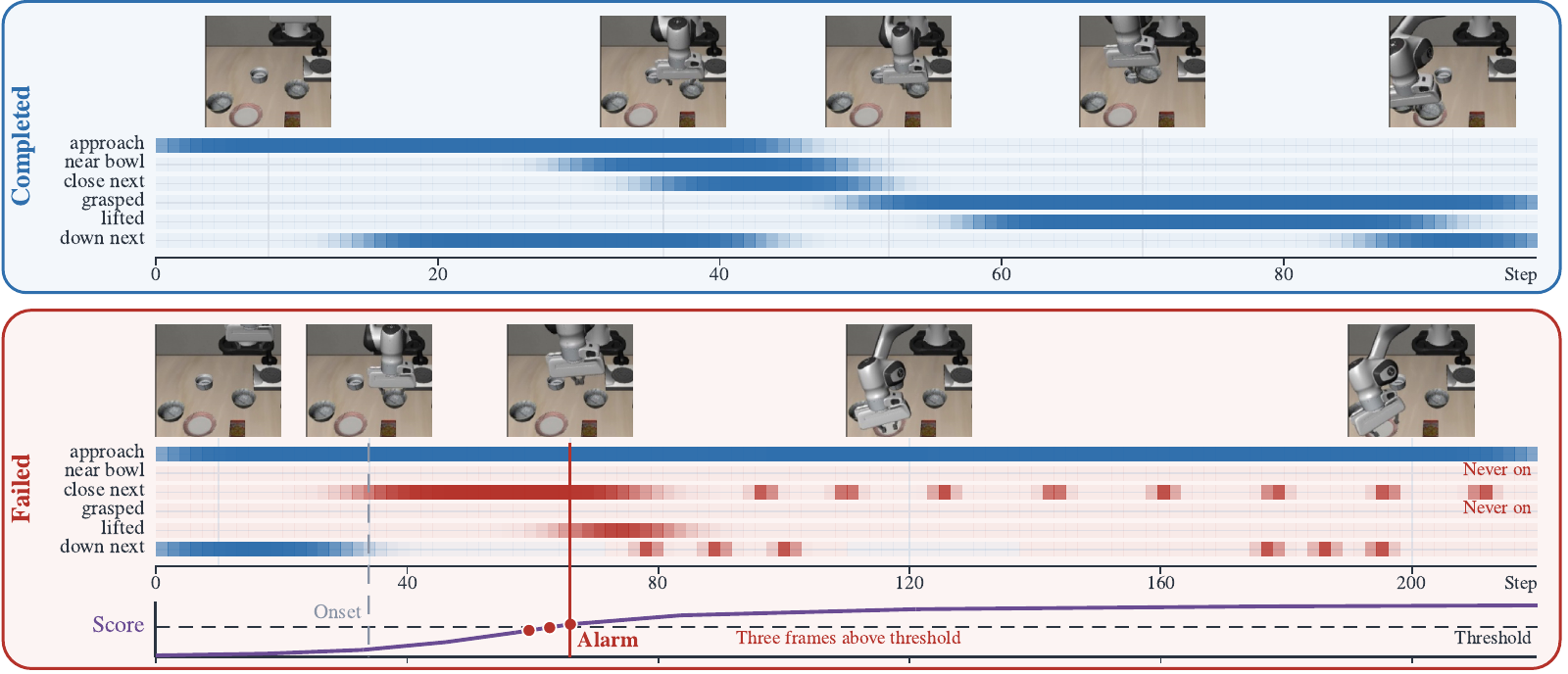} \caption{Two recorded attempts at one task: frames at the marked steps and, beneath, the six concepts
of the toy vector, shaded by how strongly each basis feature is on (timing recorded, shading
illustrative). Top: success. Bottom: failure, where the policy represents the gripper closing early,
never represents the gripper near the bowl or the bowl grasped, and switches the downward decision on
and off. The alarm is the first step at which the score stays above the threshold for three frames.}
\label{fig:course}
\end{figure}

At every step the monitor decides from the named basis features up to step $t$ whether the episode is
going wrong, not from their names.
Its two parts are a record $M$ and an alarm $A$.
For each named basis feature $i$ at step $t$ the record keeps five numbers.
The first is the activity $b_{t,i} \in \{0,1\}$, which is one when the basis feature is on, with
$\alpha_i > 0$, at any position of step $t$.
Next come the three dwells $\mu^{j}_{t,i}$, running means of $b_{t,i}$ over a short, a middle and a
long window of $k_j$ frames, whose lengths are fixed before the alarm is fitted
(Appendix~\ref{app:monitor}).
Last comes the switching rate, a running mean of $\lvert b_{t,i} - b_{t-1,i}\rvert$ over a fixed
window, high when a basis feature turns on and off instead of settling.
On the failed episode of Figure~\ref{fig:course} the dwells of $g_4$ stay at zero, while the
switching rate of $g_6$ climbs as the arm hovers above the table.

Gathered over the named basis features, the five numbers form one vector $z_t$, which the alarm scores as
$s_t = w^{\top} z_t + w_0$ with a logistic model.
We fit the weights $w$ on the fitting episodes, with the frames of the successful ones as the
negative class.
An episode fails when it never reaches the last \emph{phase} of the task, a stage such as the bowl
grasped that the simulator or, on the real robot, the automatic labels of Section~\ref{sec:setup}
mark.
Its frames form the positive class from its \textbf{failure onset}, the earlier of the last step at
which it drops the object and the step by which a fixed share of the successful fitting episodes had
reached the first phase that it never reaches.
Earlier frames of a failed episode, where the policy may still be doing what a successful episode
does, stay out of the fit.
For the alarm threshold we take the score at which a fixed small fraction of the successful fitting
episodes produce a false alarm.
The monitor raises an alarm the first time its score stays above that threshold for a fixed number of
consecutive frames, which keeps a single noisy frame from calling a failure
(Appendix~\ref{app:monitor} lists these constants).

\subsection{The explanation of a failure}
\label{sec:method:cause}

Because the score is linear in $z_t$, it splits into one term per coordinate of $z_t$ with no
approximation,
\begin{equation}
s_t - \bar{s} \;=\; \sum_{j} w_j \bigl( z_{t,j} - \bar{z}_j \bigr),
\qquad
\bar{s} \;=\; w^{\top} \bar{z} + w_0 ,
\label{eq:contrib}
\end{equation}
where $\bar{z}$ is the mean of $z_t$ over the successful frames of the fitting episodes and $\bar{s}$
the score of an average successful frame.
Term $j$ of Equation~\ref{eq:contrib} measures how far one coordinate has drifted from its usual value
on a successful episode and how much that drift pushes the score toward failure.
Summed over the five numbers of every basis feature named for a concept, these terms give the
\textbf{contribution} of that concept.
Named basis features that are on in every ordinary frame are summed into one further line beside the
concepts.
Our explanation $X$ ranks the concepts by contribution, and the largest are the \textbf{responsible
concepts} of the alarm.
On the failed episode the two largest are $g_3$, which was on while $g_2$ was still off, and $g_4$,
whose dwells never left zero.
Together they report an early closure decision: the policy decided to close the gripper before it
represented the bowl as within reach, and never afterwards represented the bowl as grasped.
A language model, named in Section~\ref{sec:setup}, writes this explanation as a short note, the
\textbf{diagnosis} of the episode.
Its input is text drawn from the monitor alone: the instruction, the step intervals in which each named
concept was on, what a successful episode would already show by that step, the alarm step and the
responsible concepts.
Its prompt allows no concept outside that list, and a concept whose basis feature never came on is reported
as something the policy did not represent, never as a fact about the bowl or the gripper.
Decoding is greedy, which makes the same input always produce the same note.
\section{Experiments}
\label{sec:experiments}
\subsection{Experimental setup}
\label{sec:setup}

\textbf{Policies and tasks:}
The experiments test three claims of Section~\ref{sec:intro}: that names stay valid on held-out
episodes, that the policy acts through the named basis features, and that these basis features detect
and explain failures.
We test them on three VLAs that produce their actions in different ways.
OpenVLA \citep{kim2024openvla} writes each action as tokens of its language model.
Its released checkpoints run the Spatial, Object and Goal suites of LIBERO \citep{liu2023libero},
each with ten tasks and twenty initial states.
LingBot-VLA~2.0 \citep{wu2026lingbot} and $\pi_{0.5}$ \citep{pi05_2025} instead add an action expert
to a vision-language backbone.
LingBot-VLA runs $200$ episodes on two RoboTwin~2.0 tasks \citep{chen2025robotwin}, lift\_pot and
click\_bell.
In LIBERO, $\pi_{0.5}$ runs one task per suite, where the policy was very accurate so we displaced and rotated objects  to make $30\%$ of
attempts fail.
On real hardware, a Franka Panda arm with an Inspire hand runs a fine-tuned $\pi_{0.5}$.
In the stove task it places a pot on a stove and then a lid on the pot, and in the cup task it stacks
two cups inside a third.
Appendix~\ref{app:setup} describes the policies and tasks in detail.

\textbf{Concepts and baselines:}
Each stream of a policy gets one dictionary with four times as many basis features as its layer is
wide.
The dictionaries cover layer $20$ of OpenVLA and layer $22$ of the LingBot-VLA action expert.
For $\pi_{0.5}$ they cover both streams, at layers $7$, $11$ and $15$ in simulation and at layer $11$
on the robot.
The basis features are named after two kinds of concepts.
State concepts describe the gripper or hand, the phase of the task, and whether an object is touched,
grasped, lifted or carried.
Decision concepts give the sign of each action axis and the gripper command.
The simulator labels every simulated frame with these concepts.
On the real robot the labels come automatically from arm kinematics, hand joint sensors and object masks in
the camera images, and they contain no error on $30$ demonstrations which were further verified by a human.
The naming baselines apply the rule of Section~\ref{sec:method:name} to raw coordinates of the
activation, to random directions and to shuffled concept labels.
On OpenVLA they also include feed-forward neurons and the dictionary of an untrained copy of the policy.
The detection baseline is a dense monitor on the full activation and its running means, with our
labels and thresholds.

\textbf{Evaluation protocol:}
Names are chosen on the fitting episodes and tested on held-out episodes.
In LIBERO the fitting episodes cover nine tasks at twelve of the twenty initial states.
The held-out episodes are the other eight initial states and a tenth task that no part of MIND has
seen.
RoboTwin keeps episodes $120$ to $199$ for testing, $\pi_{0.5}$ in LIBERO is tested on new initial
states, and the robot on five attempts per task.
A name counts only if it passes the rule of Section~\ref{sec:method:name} again on these episodes, with
precision at least $0.9$.
For detection we report the AUROC over held-out episodes with a $95\%$ bootstrap interval, and the
share of successful episodes with a false alarm.
We score each robot attempt with monitors fitted on all the others, since the attempts are too few for
a separate test set.
Because a failed RoboTwin episode runs to the step limit, its monitors are scored on the opening
frames, where length alone gives an AUROC of $0.50$.
Qwen2.5-7B-Instruct \citep{yang2024qwen25} writes the note of Section~\ref{sec:method:cause} with
greedy decoding.

\subsection{Simulation experiments}
\label{sec:results}
\label{sec:results:sim}

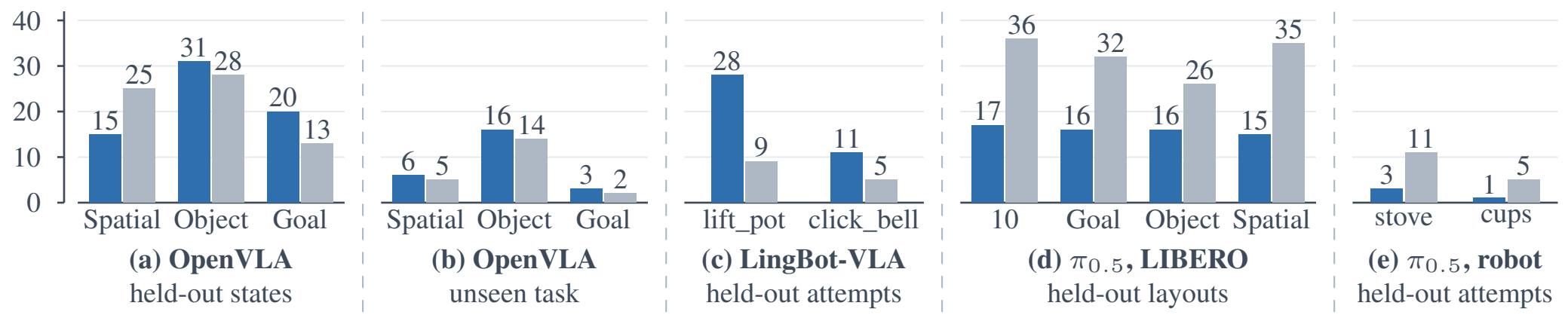
\begin{figure}[t]
\centering
\begin{tikzpicture}[x=1cm, y=0.04cm, font=\scriptsize]
  \foreach \yy in {10,20,30,40}{\draw[mindoff, line width=0.4pt] (0.120,\yy) -- (2.460,\yy);}
  \draw[mindink, line width=0.5pt] (0.120,0) -- (2.460,0);
  \fill[mindon] (0.220,0) rectangle (0.500,15); \fill[mindfaint] (0.520,0) rectangle (0.800,25);
  \node[mindink, anchor=south, inner sep=1pt] at (0.360,15) {15}; \node[mindink, anchor=south, inner sep=1pt] at (0.660,25) {25};
  \node[mindink, anchor=north, inner sep=1.5pt] at (0.510,0) {Spatial};
  \fill[mindon] (1.000,0) rectangle (1.280,31); \fill[mindfaint] (1.300,0) rectangle (1.580,28);
  \node[mindink, anchor=south, inner sep=1pt] at (1.140,31) {31}; \node[mindink, anchor=south, inner sep=1pt] at (1.440,28) {28};
  \node[mindink, anchor=north, inner sep=1.5pt] at (1.290,0) {Object};
  \fill[mindon] (1.780,0) rectangle (2.060,20); \fill[mindfaint] (2.080,0) rectangle (2.360,13);
  \node[mindink, anchor=south, inner sep=1pt] at (1.920,20) {20}; \node[mindink, anchor=south, inner sep=1pt] at (2.220,13) {13};
  \node[mindink, anchor=north, inner sep=1.5pt] at (2.070,0) {Goal};
  \node[mindink, anchor=north, inner sep=0pt, font=\scriptsize\bfseries] at (1.290,-10) {(a) OpenVLA};
  \node[mindink, anchor=north, inner sep=0pt] at (1.290,-17.5) {held-out states};
  \foreach \yy in {10,20,30,40}{\draw[mindoff, line width=0.4pt] (2.780,\yy) -- (5.120,\yy);}
  \draw[mindink, line width=0.5pt] (2.780,0) -- (5.120,0);
  \fill[mindon] (2.880,0) rectangle (3.160,6); \fill[mindfaint] (3.180,0) rectangle (3.460,5);
  \node[mindink, anchor=south, inner sep=1pt] at (3.020,6) {6}; \node[mindink, anchor=south, inner sep=1pt] at (3.320,5) {5};
  \node[mindink, anchor=north, inner sep=1.5pt] at (3.170,0) {Spatial};
  \fill[mindon] (3.660,0) rectangle (3.940,16); \fill[mindfaint] (3.960,0) rectangle (4.240,14);
  \node[mindink, anchor=south, inner sep=1pt] at (3.800,16) {16}; \node[mindink, anchor=south, inner sep=1pt] at (4.100,14) {14};
  \node[mindink, anchor=north, inner sep=1.5pt] at (3.950,0) {Object};
  \fill[mindon] (4.440,0) rectangle (4.720,3); \fill[mindfaint] (4.740,0) rectangle (5.020,2);
  \node[mindink, anchor=south, inner sep=1pt] at (4.580,3) {3}; \node[mindink, anchor=south, inner sep=1pt] at (4.880,2) {2};
  \node[mindink, anchor=north, inner sep=1.5pt] at (4.730,0) {Goal};
  \node[mindink, anchor=north, inner sep=0pt, font=\scriptsize\bfseries] at (3.950,-10) {(b) OpenVLA};
  \node[mindink, anchor=north, inner sep=0pt] at (3.950,-17.5) {unseen task};
  \foreach \yy in {10,20,30,40}{\draw[mindoff, line width=0.4pt] (5.440,\yy) -- (7.540,\yy);}
  \draw[mindink, line width=0.5pt] (5.440,0) -- (7.540,0);
  \fill[mindon] (5.675,0) rectangle (5.955,28); \fill[mindfaint] (5.975,0) rectangle (6.255,9);
  \node[mindink, anchor=south, inner sep=1pt] at (5.815,28) {28}; \node[mindink, anchor=south, inner sep=1pt] at (6.115,9) {9};
  \node[mindink, anchor=north, inner sep=1.5pt] at (5.965,0) {lift\_pot};
  \fill[mindon] (6.725,0) rectangle (7.005,11); \fill[mindfaint] (7.025,0) rectangle (7.305,5);
  \node[mindink, anchor=south, inner sep=1pt] at (6.865,11) {11}; \node[mindink, anchor=south, inner sep=1pt] at (7.165,5) {5};
  \node[mindink, anchor=north, inner sep=1.5pt] at (7.015,0) {click\_bell};
  \node[mindink, anchor=north, inner sep=0pt, font=\scriptsize\bfseries] at (6.490,-10) {(c) LingBot-VLA};
  \node[mindink, anchor=north, inner sep=0pt] at (6.490,-17.5) {held-out attempts};
  \foreach \yy in {10,20,30,40}{\draw[mindoff, line width=0.4pt] (7.860,\yy) -- (10.980,\yy);}
  \draw[mindink, line width=0.5pt] (7.860,0) -- (10.980,0);
  \fill[mindon] (7.960,0) rectangle (8.240,17); \fill[mindfaint] (8.260,0) rectangle (8.540,36);
  \node[mindink, anchor=south, inner sep=1pt] at (8.100,17) {17}; \node[mindink, anchor=south, inner sep=1pt] at (8.400,36) {36};
  \node[mindink, anchor=north, inner sep=1.5pt] at (8.250,0) {10};
  \fill[mindon] (8.740,0) rectangle (9.020,16); \fill[mindfaint] (9.040,0) rectangle (9.320,32);
  \node[mindink, anchor=south, inner sep=1pt] at (8.880,16) {16}; \node[mindink, anchor=south, inner sep=1pt] at (9.180,32) {32};
  \node[mindink, anchor=north, inner sep=1.5pt] at (9.030,0) {Goal};
  \fill[mindon] (9.520,0) rectangle (9.800,16); \fill[mindfaint] (9.820,0) rectangle (10.100,26);
  \node[mindink, anchor=south, inner sep=1pt] at (9.660,16) {16}; \node[mindink, anchor=south, inner sep=1pt] at (9.960,26) {26};
  \node[mindink, anchor=north, inner sep=1.5pt] at (9.810,0) {Object};
  \fill[mindon] (10.300,0) rectangle (10.580,15); \fill[mindfaint] (10.600,0) rectangle (10.880,35);
  \node[mindink, anchor=south, inner sep=1pt] at (10.440,15) {15}; \node[mindink, anchor=south, inner sep=1pt] at (10.740,35) {35};
  \node[mindink, anchor=north, inner sep=1.5pt] at (10.590,0) {Spatial};
  \node[mindink, anchor=north, inner sep=0pt, font=\scriptsize\bfseries] at (9.420,-10) {(d) $\pi_{0.5}$, LIBERO};
  \node[mindink, anchor=north, inner sep=0pt] at (9.420,-17.5) {held-out layouts};
  \foreach \yy in {10,20,30,40}{\draw[mindoff, line width=0.4pt] (11.300,\yy) -- (13.100,\yy);}
  \draw[mindink, line width=0.5pt] (11.300,0) -- (13.100,0);
  \fill[mindon] (11.460,0) rectangle (11.740,3); \fill[mindfaint] (11.760,0) rectangle (12.040,11);
  \node[mindink, anchor=south, inner sep=1pt] at (11.600,3) {3}; \node[mindink, anchor=south, inner sep=1pt] at (11.900,11) {11};
  \node[mindink, anchor=north, inner sep=1.5pt] at (11.750,0) {stove};
  \fill[mindon] (12.360,0) rectangle (12.640,1); \fill[mindfaint] (12.660,0) rectangle (12.940,5);
  \node[mindink, anchor=south, inner sep=1pt] at (12.500,1) {1}; \node[mindink, anchor=south, inner sep=1pt] at (12.800,5) {5};
  \node[mindink, anchor=north, inner sep=1.5pt] at (12.650,0) {cups};
  \node[mindink, anchor=north, inner sep=0pt, font=\scriptsize\bfseries] at (12.200,-10) {(e) $\pi_{0.5}$, robot};
  \node[mindink, anchor=north, inner sep=0pt] at (12.200,-17.5) {held-out attempts};
  \draw[mindfaint, line width=0.4pt, dashed] (2.620,-24) -- (2.620,44);
  \draw[mindfaint, line width=0.4pt, dashed] (5.280,-24) -- (5.280,44);
  \draw[mindfaint, line width=0.4pt, dashed] (7.700,-24) -- (7.700,44);
  \draw[mindfaint, line width=0.4pt, dashed] (11.140,-24) -- (11.140,44);
  \draw[mindink, line width=0.5pt] (0,0) -- (0,42);
  \foreach \yy in {0,10,20,30,40}{\draw[mindink, line width=0.5pt] (0,\yy) -- (-0.06,\yy); \node[mindink, anchor=east] at (-0.08,\yy) {\yy};}
\end{tikzpicture}
\caption{Names that stay precise on held-out episodes, for decision (dark) and state (light) concepts.
(a, b) OpenVLA at unseen initial states and on the unseen task. (c) LingBot-VLA, action expert, layer
$22$. (d) $\pi_{0.5}$ in LIBERO, both streams, layers $7$, $11$ and $15$ (10 is LIBERO-10). (e) $\pi_{0.5}$ on
the robot, backbone, layer $11$. No baseline names any concept in (a), (b) or (e); Tables~\ref{tab:names_lingbot}--\ref{tab:names_robot} give every baseline.}
\label{fig:names}
\end{figure}

\begin{figure}[t]
\centering
\includegraphics[width=\textwidth]{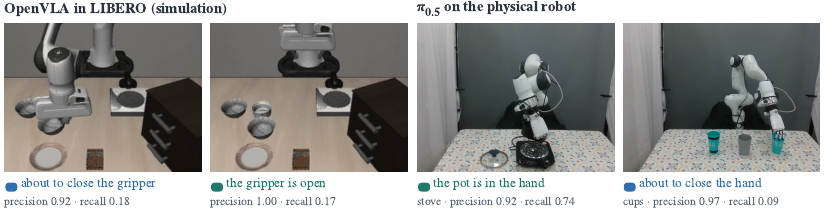}\\[5pt]
\includegraphics[width=\textwidth]{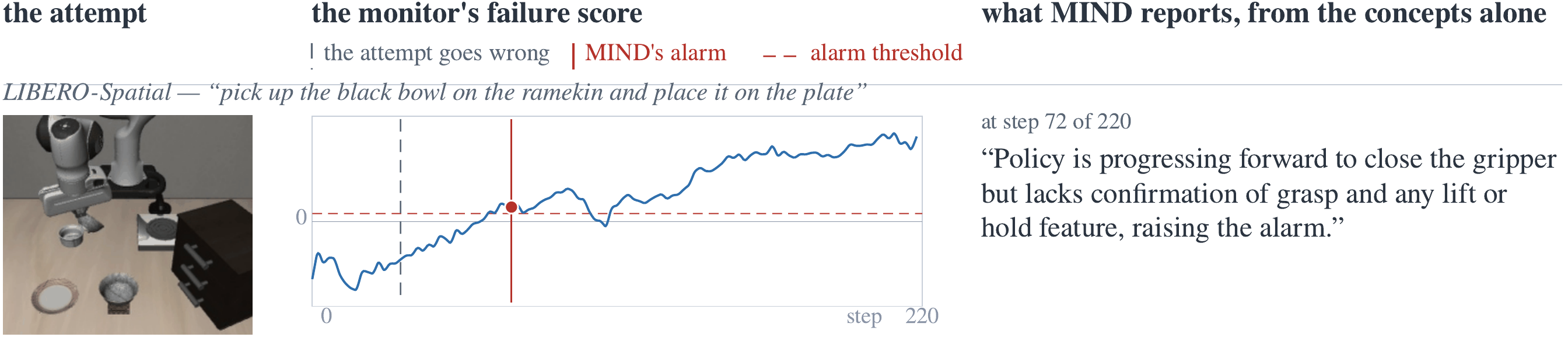}
\caption{(a) Frames at which named basis features fire hardest on held-out episodes, with the held-out
precision and recall of each name (blue: decision; green: state). (b) An OpenVLA failure:
the frame at the alarm, the failure score and the note written from the concept record alone.}
\label{fig:glimpse}
\end{figure}

\textbf{Naming:}
Names chosen by MIND stay precise on held-out episodes on every policy, and no baseline matches them
(Figure~\ref{fig:names}).
The test is strictest on OpenVLA, where a name must hold at unseen initial states and on a task that no
part of MIND has seen.
The dictionary keeps names for decisions and states there, while no baseline, neurons and the untrained
copy included, names a single concept.
The two policies with an action expert show where the advantage of the dictionary lies.
Many directions of an action expert already follow the action it writes, so raw coordinates also name
some motion concepts on LingBot-VLA and $\pi_{0.5}$.
The state of the robot is spread across coordinates, and the dictionary separates it far better than
raw coordinates do, which is where its lead is widest (Appendix~\ref{app:res:names}).
These names do not depend on the random start of the fit, since two dictionaries of OpenVLA fitted from
different starts share $81$ to $83\%$ of their named basis features (Appendix~\ref{app:res:sparsity}).
Each name also marks a moment that a person can check in the recording, as
Figure~\ref{fig:glimpse}(a) shows, where the basis feature for closing the gripper peaks as the gripper
reaches the bowl.
\textbf{Steering:}
Clamping a named basis feature in closed loop shows that the policy acts through it.
On OpenVLA the clamp drives the arm the way the concept says for twelve of the fourteen motion
concepts, raising the share of such steps from at most $0.25$ to at least $0.92$, while other basis
features clamped to the same multiple and random directions barely move it
(Appendix~\ref{app:res:steer}).
On $\pi_{0.5}$ the clamp moves the arm just as reliably along the named axis, but against the named
sign, since the basis feature named for $+x$ lowers the share of steps along $+x$ from $0.36$ to $0.06$,
against $0.48$ for random basis features.
This is not an error of the name.
The gradient of the action points against every direction that predicts the same decision, supervised
probes included, so the inversion lies in how $\pi_{0.5}$ encodes its decisions, and the name still
identifies the quantity through which the policy acts.
\textbf{Detection:}
The named basis features detect failures as well as the dense activation, although every input of
their monitor carries a name (Table~\ref{tab:detect}).
On OpenVLA and $\pi_{0.5}$ the two monitors tie within their intervals, and the fused variant chosen
before any fit is the best on Object and Goal.
The named monitor gains where a deployed monitor needs it.
On $\pi_{0.5}$ it catches $99\%$ of the failures against $95\%$ at half the false alarms, a median of
$10$ steps after the failure onset against $25$.
On OpenVLA only $7$ to $21\%$ of the alarms of its fused variant precede the onset, against $29$ to
$38\%$ of the dense alarms on Spatial and Object, so its alarms respond to the failure and not to early differences between episodes.
On LingBot-VLA the result depends on when a task fails.
Both monitors see only the opening frames of a RoboTwin episode, because a failed one runs to the
step limit.
On click\_bell a failing arm never reaches the bell, the failure shows in those frames, and the named
basis features lead.
On lift\_pot most failures come after both arms grasp the pot, beyond the scored frames, so neither
monitor can see them (Appendix~\ref{app:res:detect}).
\textbf{Diagnosis:}
The same basis features explain the failures they detect in the policy's own terms.
In Figure~\ref{fig:glimpse}(b) the policy decides to close the gripper while no basis feature for a
grasp or a lift turns on, and the note reports this missing grasp.
These explanations are stable, since the top responsible concept of an alarm stays among the three
largest in a median of $90$ to $95\%$ of twenty bootstrap refits of the monitor.
They are also faithful to the record, since on OpenVLA $85$ to $89\%$ of the model's answers about the grasp,
the hold, the placing and the motion match it, and $11$ of $12$ successful attempts are called normal.
The type of a failure is a separate question, bounded by the concepts that the record contains.
Against the simulator the category of the note is right for $15$ of $48$ OpenVLA failures and a third of
those of $\pi_{0.5}$, about as often as a mechanical rule that reads the same record.
The limit therefore lies in the phases that the concepts cover, not in the explanation
(Appendix~\ref{app:res:diag}).

\begin{table}[t]
\centering
\caption{Failure detection: AUROC over held-out episodes, with false alarms (share of successful
episodes) in brackets. Dense: monitor on the full activation; Ours: on the named basis features;
$^\dagger$best variant fixed before any fit. The robot is scored leaving one attempt out.
Table~\ref{tab:detect_full} gives intervals and variants; best AUROC in bold.}
\label{tab:detect}
\small
\setlength{\tabcolsep}{2.6pt}
\begin{tabular}{@{}lcccccccc@{}}
\toprule
& \multicolumn{3}{c}{OpenVLA} & \multicolumn{2}{c}{LingBot-VLA} & $\pi_{0.5}$ & \multicolumn{2}{c}{$\pi_{0.5}$, real robot} \\
\cmidrule(lr){2-4}\cmidrule(lr){5-6}\cmidrule(lr){7-7}\cmidrule(l){8-9}
Monitor & Spatial & Object & Goal & lift\_pot & click\_bell & 4 tasks & stove & cups \\
\midrule
Dense & \textbf{1.00} (2\%) & 0.91 (7\%) & 0.92 (7\%) & \textbf{0.60} (4\%) & 0.63 (3\%) & 0.988 (6\%) & 0.74 (13\%) & 0.61 (13\%) \\
Ours & \textbf{1.00} (3\%) & 0.89 (5\%) & 0.92 (5\%) & 0.58 (4\%) & \textbf{0.72} (2\%) & \textbf{0.991} (3\%) & 0.70 (13\%) & 0.60 (0\%) \\
Ours$^\dagger$ & \textbf{1.00} (2\%) & \textbf{0.93} (9\%) & \textbf{0.97} (2\%) & -- & -- & -- & \textbf{0.84} (13\%) & \textbf{0.70} (13\%) \\
\bottomrule
\end{tabular}
\end{table}

\subsection{Real-world experiments}
\label{sec:results:real}

On the physical robot the backbone of $\pi_{0.5}$ names states and decisions about the coming action
that no baseline names (Figures~\ref{fig:names}(e) and~\ref{fig:glimpse}(a)).
A linear probe finds every concept in both streams on held-out attempts, so the policy represents all
of them (Appendix~\ref{app:res:names}).
Detection on the robot is limited by what a failed attempt records, not by the monitor.
The operator stops a failing attempt about two seconds after it goes wrong, so only $9$ to $14\%$ of its
frames show the failure, against $64$ to $66\%$ on OpenVLA.
The main monitors, on the named basis features and on the dense activation alike, therefore stay within
intervals that include chance (Table~\ref{tab:detect_full}).
More attempts would not change this, since at the robot's counts the named basis features of OpenVLA
still reach a median AUROC of $0.86$ to $1.00$.
Even from these short records, the best monitor fixed before any fit catches $8$ of the $11$ stove
failures with one false alarm, a median of two seconds before the operator stops the arm.
On the cups it catches every attempt that failed on the second cup and none of those that failed on
the first, which stop before any state that the monitor learned (Appendix~\ref{app:res:detect}).
The note names the right failure for $3$ of the $8$ stove failures that the labels can score.
Its main confusion, a lid never grasped against a lid not placed, comes from basis features for holding
the lid that fire briefly in attempts that never grasped it (Appendix~\ref{app:res:diag}).

\subsection{Ablations}
\label{sec:results:ablation}

In each ablation we replace one part of MIND and keep the rest unchanged (Table~\ref{tab:ablation} in
Appendix~\ref{app:res:ablation}).
Training produces the structure that MIND names.
The dictionary of an untrained OpenVLA names no concept, and on the robot a probe on the action expert of
an untrained $\pi_{0.5}$ falls to chance.
The dictionary separates what neurons mix.
Feed-forward neurons name no concept on OpenVLA, and on $\pi_{0.5}$ they miss the decisions tied to one
step of the chunk, naming one where the dictionary names six.
Memory matters for detection, since without the dwells and the switching rate the monitor loses $0.08$
to $0.13$ AUROC on Object and Goal.
Finally, the monitor does not merely time the attempt.
A rule that alarms once an attempt outlasts $95\%$ of the successful ones fires $49$ to $158$ steps after
the failure onset, against $10$ to $28$ for the named basis features.

\section{Conclusion}
\label{sec:conclusion}
\label{sec:limitations}
MIND describes what a VLA computes through sparse basis features whose names are tested on unseen
episodes.
Across OpenVLA, LingBot-VLA and $\pi_{0.5}$ in simulation and $\pi_{0.5}$ on a physical robot, the
named basis features cover concepts that raw coordinates, random directions and shuffled labels miss,
and clamping one of them steers the arm in closed loop.
The same basis features detect failures as well as the dense activation, with fewer false alarms on
most tasks, and their record lets a language model describe each failure in the policy's own concepts.
Next, a closed-loop check on the robot will carry the clamp beyond simulation, richer phase concepts
will let the note separate the failure types that the current concepts do not yet distinguish, and
robot attempts with varied object placements and full-length failures will give the monitor more of
each failure to learn from.

\section*{Reproducibility statement}
Section~\ref{sec:method} defines every part of MIND, and Appendices~\ref{app:sae} to~\ref{app:name}
give every constant that it uses, all fixed before any result was examined: the fit of the
dictionaries in Appendix~\ref{app:sae}, and the constants of the monitor and of the naming rule in
Tables~\ref{tab:monitor} and~\ref{tab:name}.
Appendix~\ref{app:setup} gives the policies, the tasks, the splits, the labels and the protocol of
every measurement, and Appendix~\ref{app:results} gives the full results with their intervals.
The anonymized supplementary material holds what is needed to regenerate them.
Its \texttt{README.md} states the protocol that every policy follows, with the rules that decide each
result.
Its \texttt{reference\_openvla/} folder is the complete implementation for OpenVLA on LIBERO, from the
collection of rollouts to the note of the language model.
It contains the tests that check the extraction of activations and the recovery of planted
directions by the sparse autoencoder, a guide that regenerates every table command by command, and
the result files from which the tables of the paper are read by script.
Its \texttt{results/lingbot\_robotwin/} folder gives the results of LingBot-VLA on RoboTwin with the
configuration, commands and hyperparameters that regenerate them, and its \texttt{videos/} folder shows
detected failures, closed-loop steering and concept timelines on held-out episodes.
The experiments on $\pi_{0.5}$ follow the same protocol and fit their dictionaries with the same code.
The policies in simulation, the simulators and the language model are public: OpenVLA, LingBot-VLA~2.0
and $\pi_{0.5}$, LIBERO and RoboTwin~2.0, and Qwen2.5-7B-Instruct with greedy decoding.

\bibliography{iclr2027_conference}
\bibliographystyle{iclr2027_conference}

\appendix
\section{Fitting the sparse autoencoder}
\label{app:sae}
The dictionary of Section~\ref{sec:method:sae} is fitted by the procedure of
\citet{templeton2024scaling}.
Every activation vector is multiplied by one constant, chosen on the fitting episodes so that the
scaled vectors have an average length of $\sqrt{d}$.
The decoder columns start as random directions of length $0.1$, the encoder starts as the transpose
of the decoder, and both biases start at zero.
The fit uses Adam without weight decay, the weight $\lambda$ rises linearly from zero over the first
five percent of the steps, the learning rate falls linearly to zero over the last twenty percent,
and the gradient norm is clipped at one.
Because the sparsity term of Equation~\ref{eq:saeloss} weights each coefficient by the length of its
column, the columns need no unit-norm constraint: shrinking $f_i$ while lengthening its column by
the same factor leaves $\alpha_i(x)$ unchanged.
The dictionary has $m = 4d$ columns, and the weight $\lambda$ is the one, among the values tried,
at which the average number of basis features on per activation vector is nearest $50$.
On OpenVLA each dictionary is fitted on three to four million vectors of its suite.
On LingBot-VLA the dictionary reads the action expert at its fifty action positions, taken at the
last denoising step, and at its state position.
Its backbone is left out, because a linear probe on it recovers only $7$ of $21$ concepts.
Dictionaries of $4d = 3{,}072$ basis features are fitted at layers $12$, $22$ and $30$ of the $36$, and
layer $22$ is reported because it names the most concepts on both tasks.
The weight is chosen per layer for about eighty basis features on, since one weight gives very different
sparsity at different layers of the action expert.
On the physical robot every vector of $\pi_{0.5}$ carries an offset shared by all vectors, $88$ to
$95\%$ of its length, and with it no weight gives a dictionary near fifty basis features on.
The mean vector of the fitting frames is therefore subtracted before the fit and from every vector
that is named or tested.
The action expert gets one mean per denoising step, since its tokens differ in size and direction
from one step to the next.
The fit sees every step of the fitting attempts, $1.73$ million backbone vectors and $489{,}500$
action-expert vectors on the stove task, the latter at all ten denoising steps; names are tested at
the last.
The rule of fifty basis features on picks $\lambda = 2$ for the backbone on both tasks, and $16$ and $12$
for the action expert on the stove and cup tasks.
On $\pi_{0.5}$ in LIBERO the weight is set per stream, $10$ for the backbone and $5$ for the action
expert: the weight nearest fifty basis features on leaves more variance unexplained than the mean vector at
four of its six layers and streams.

\section{Constants of the monitor}
\label{app:monitor}
Table~\ref{tab:monitor} lists the constants of the record and the alarm of
Section~\ref{sec:method:monitor}.
They were fixed before the alarm was fitted.
On LingBot-VLA, whose frame spans ten steps of the robot, the dwell windows are $\{1, 4, 16\}$
frames.
\begin{table}[H]
\centering
\caption{Constants of the record and the alarm.}
\label{tab:monitor}
\begin{tabular}{lc}
\toprule
meaning & value \\
\midrule
dwell window lengths $k$, in frames & $\{4, 16, 64\}$ \\
switching-rate window, in frames & $16$ \\
share of successful episodes that fixes the onset deadline & $0.9$ \\
false-alarm fraction on successful fitting episodes that sets the threshold & $0.05$ \\
consecutive frames above the threshold before an alarm & $3$ \\
\bottomrule
\end{tabular}
\end{table}

\section{Constants of the naming rule}
\label{app:name}
Table~\ref{tab:name} lists the constants of the naming rule, of the held-out test and of the
check of Section~\ref{sec:method:name}.
They were fixed before any basis feature was examined and are the same for every policy, for every
control, and for the naming rule and its held-out test.
\begin{table}[H]
\centering
\caption{Constants of the naming rule.}
\label{tab:name}
\begin{tabular}{llc}
\toprule
symbol & meaning & value \\
\midrule
$\mathcal{T}$ & firing levels tried, as fractions of $\alpha^{\max}_i$ & $\{0, 0.1, 0.2, 0.3\}$ \\
$p_{\min}$ & precision floor & $0.9$ \\
$r_{\min}$ & recall floor of a named basis feature & $0.05$ \\
$n_{\min}$ & minimum number of firings & $50$ \\
-- & minimum number of episodes a holding name fires in & $5$ \\
-- & largest share of those firings from one episode & $0.5$ \\
-- & clamp, as a multiple of $\alpha^{\max}_i$ & $1.5$ \\
\bottomrule
\end{tabular}
\end{table}

\begin{figure}[H]
\begin{minipage}[c]{0.46\textwidth}
\centering
\scalebox{0.94}{%
\begin{tikzpicture}[x=1mm,y=1mm]

\node[mtitle, anchor=west] at (0,70.5) {(a) naming frames};

\foreach \k/\h/\c in {0/1.2/mindfaint, 1/0.9/mindfaint, 2/7.0/mindon, 3/9.5/mindon,
   4/11.0/mindon, 5/7.8/mindon, 6/2.2/mindfaint, 7/8.2/mindbad, 8/1.0/mindfaint,
   9/6.8/mindon, 10/10.2/mindon, 11/7.4/mindon, 12/1.6/mindfaint, 13/1.5/mindfaint,
   14/9.0/mindon, 15/3.2/mindfaint}{
  \fill[\c] (10+\k*2.6,55) rectangle (12.3+\k*2.6,55+\h);
}
\draw[mthin] (10,55) -- (51.3,55);
\draw[dashed, line width=0.4pt, draw=mindink] (9.6,60) -- (51.7,60);
\node[mnote, anchor=east] at (9.0,66.0) {$\alpha_3(x)$};
\node[mnote, anchor=east] at (9.0,60.0) {$\tau\,\alpha^{\max}_3$};

\foreach \k in {0,...,15}{
  \fill[mindoff] (10+\k*2.6,50.6) rectangle (12.3+\k*2.6,53.4);
  \draw[mrule]   (10+\k*2.6,50.6) rectangle (12.3+\k*2.6,53.4);
}
\foreach \k in {2,3,4,5,9,10,11,14,15}{
  \fill[mindon] (10+\k*2.6,50.6) rectangle (12.3+\k*2.6,53.4);
  \draw[mrule]  (10+\k*2.6,50.6) rectangle (12.3+\k*2.6,53.4);
}

\draw[line width=0.5pt, draw=mindbad] (28.45,48.4) -- (30.25,50.2);
\draw[line width=0.5pt, draw=mindbad] (28.45,50.2) -- (30.25,48.4);
\draw[line width=0.5pt, draw=mindbad] (50.15,49.3) circle (0.9);

\node[mnote, anchor=west] at (52.6,56.6) {$P=8/9$};
\node[mnote, anchor=west] at (52.6,52.0) {$R=8/9$};
\node[mnote, anchor=west] at (10.0,46.0) {gripper closes next};

\draw[marrow] (30.65,44.4) -- (30.65,41.2);
\node[mnote, anchor=west] at (31.6,42.8) {choose};

\node[mtitle, anchor=west] at (0,39.0) {(b) candidates for $g_3$};

\node[mnote, anchor=west] at (0,33.1) {gripper closes next};
\fill[mindon]  (25.0,33.2) rectangle (50.4,34.6);
\fill[mindon]  (25.0,31.6) rectangle (49.8,32.6);
\node[mnote, anchor=east] at (24.2,33.9) {$P$};
\node[mnote, anchor=east] at (24.2,32.1) {$R$};
\draw[line width=0.6pt, draw=mindon] (52.2,33.6) -- (53.2,32.6) -- (55.0,35.0);

\node[mnote, anchor=west] at (0,27.8) {near the bowl};
\fill[mindfaint] (25.0,27.1) rectangle (40.2,28.5);
\node[mnote, anchor=west] at (0,24.1) {bowl grasped};
\fill[mindfaint] (25.0,23.4) rectangle (30.2,24.8);

\draw[dashed, line width=0.4pt, draw=mindink] (47.0,22.8) -- (47.0,35.6);
\node[mnote, anchor=south] at (47.0,35.9) {$p_{\min}$};

\draw[marrow] (30.65,21.9) -- (30.65,18.7);
\node[mnote, anchor=west] at (31.6,20.3) {then test};

\node[mtitle, anchor=west] at (0,16.5) {(c) held-out frames};
\draw[line width=0.6pt, draw=mindon] (43.0,16.3) -- (44.0,15.3) -- (45.8,17.7);
\node[mnote, anchor=west] at (46.4,16.5) {holds};

\foreach \k/\h/\c in {0/1.0/mindfaint, 1/7.2/mindon, 2/10.0/mindon, 3/8.0/mindon,
   4/1.4/mindfaint, 5/0.8/mindfaint, 6/2.0/mindfaint, 7/6.9/mindon, 8/9.4/mindon,
   9/7.6/mindon, 10/1.1/mindfaint, 11/1.7/mindfaint, 12/8.6/mindon, 13/10.6/mindon,
   14/6.5/mindon, 15/0.9/mindfaint}{
  \fill[\c] (10+\k*2.6,3.0) rectangle (12.3+\k*2.6,3.0+\h);
}
\draw[mthin] (10,3.0) -- (51.3,3.0);
\draw[dashed, line width=0.4pt, draw=mindink] (9.6,8.0) -- (51.7,8.0);

\foreach \k in {0,...,15}{
  \fill[mindoff] (10+\k*2.6,-1.4) rectangle (12.3+\k*2.6,1.4);
  \draw[mrule]   (10+\k*2.6,-1.4) rectangle (12.3+\k*2.6,1.4);
}
\foreach \k in {1,2,3,7,8,9,12,13,14}{
  \fill[mindon] (10+\k*2.6,-1.4) rectangle (12.3+\k*2.6,1.4);
  \draw[mrule]  (10+\k*2.6,-1.4) rectangle (12.3+\k*2.6,1.4);
}
\node[mnote, anchor=west] at (52.6,4.6) {$P=9/9$};
\node[mnote, anchor=west] at (52.6,0.0) {$R=9/9$};

\end{tikzpicture}}
\end{minipage}\hfill
\begin{minipage}[c]{0.50\textwidth}
\caption{How $g_3$ of the toy vector is named by the rule of Section~\ref{sec:method:name}.
(a) Naming frames.
Each bar is the activation $\alpha_3$ on one frame, the dashed line is the firing level
$\tau\,\alpha^{\max}_3$, and the basis feature fires where a bar rises above it.
The row beneath marks the frames on which the concept, the gripper closing next, is true.
The red bar is a firing on a frame where the concept is false, and it lowers the precision $P$.
The circled frame is one where the concept is true but the basis feature is off, and it lowers the
recall $R$.
(b) Candidates.
Of the concepts $g_3$ fires with, only the gripper closing next clears the precision floor
$p_{\min}$, and it becomes the name.
(c) Held-out frames.
Precision and recall are measured again with the name and its level fixed, and the name holds.}
\label{fig:name}
\end{minipage}
\end{figure}
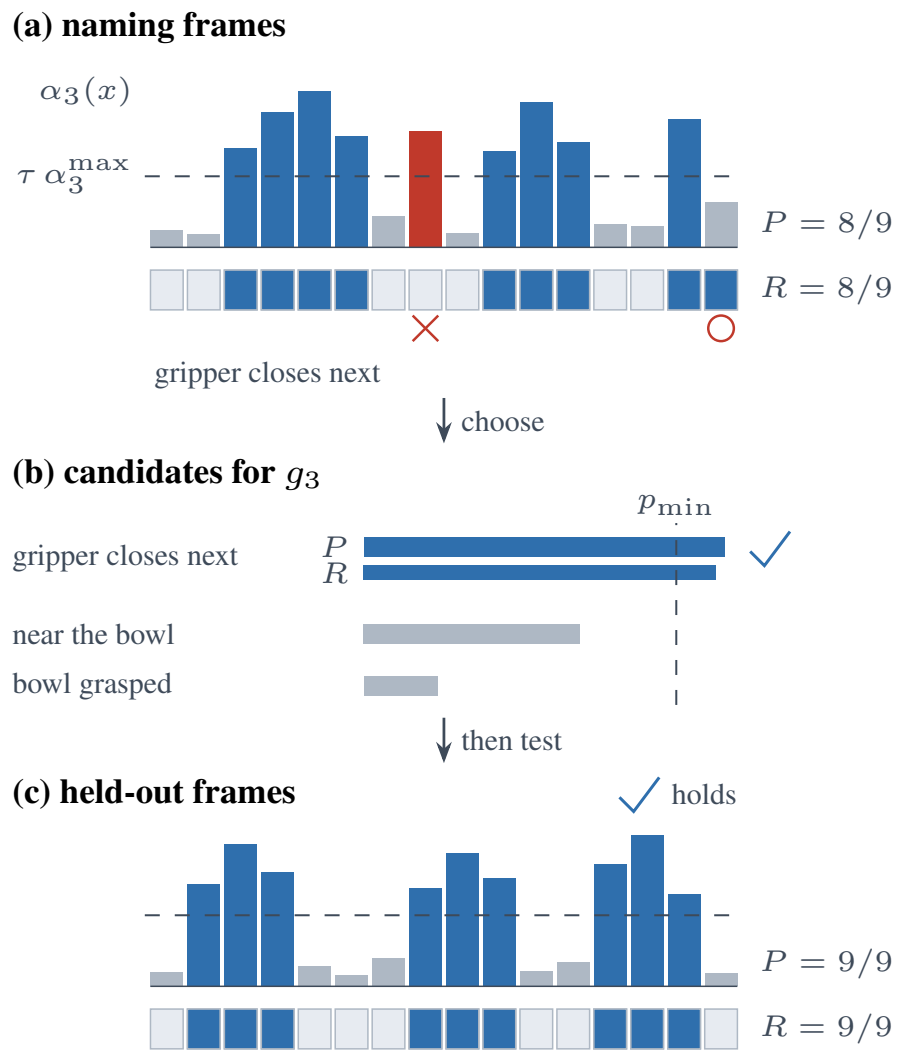

\section{The experimental protocol in detail}
\label{app:setup}
\textbf{Policies and tasks in detail.}
OpenVLA has seven billion parameters.
LingBot-VLA~2.0 has a backbone of four billion parameters, and its action expert returns fifty
actions, of which the robot executes ten.
In lift\_pot both arms lift a pot under a changing camera distance and lighting, and $58$ of its $91$
failures come after both arms have grasped the pot.
In click\_bell one arm presses a bell, and all $24$ of its failures end short of the bell.
$\pi_{0.5}$ has a backbone of two billion parameters, and its action expert returns a chunk of ten
actions by flow matching.
Its released LIBERO checkpoint runs one task of each of the four suites, including the long-horizon
LIBERO-10.
Every movable object is displaced by up to $0.12$ to $0.13$\,m and rotated by up to $\pm 45^\circ$,
which makes $120$ of the first $400$ attempts fail, and $78$ further failures are recorded for the
monitor.
On the physical robot the Franka Panda arm has a six-joint Inspire hand and works at $20$ steps a
second under one fixed RGB-D camera, with a fine-tuned checkpoint per task.
The robot fails $11$ of $19$ stove attempts and $14$ of $22$ cup attempts.
All $11$ stove failures happen on the lid, and all $14$ cup failures come from one arrangement of the
objects.

\textbf{Dictionaries and concepts in detail.}
The dictionary has $m = 16{,}384$ basis features at layer $20$ of OpenVLA, where $d = 4096$, and
$m = 3{,}072$ at layer $22$ of the action expert of LingBot-VLA, where $d = 768$.
On $\pi_{0.5}$ it has $m = 4d$ basis features on the backbone, where $d = 2048$, and on the action expert,
where $d = 1024$.
On the robot the mean vector of the fitting frames is subtracted first, one mean per denoising step for
the action expert (Appendix~\ref{app:sae}).
On $\pi_{0.5}$ each decision concept is also taken at each of the ten steps of the chunk, and on
OpenVLA and $\pi_{0.5}$ in LIBERO object concepts name the object in each image patch.
On the robot the object masks come from a text-prompted detector \citep{liu2024grounding} and a video
segmentation model \citep{ravi2024sam2}, with the depth image.
The clamp runs OpenVLA in closed loop on the unseen task, from six to eight initial states per suite.

On LIBERO the fitting episodes are the first nine tasks of a suite at their first twelve initial
states, and on RoboTwin the even episodes among the first $120$ fit the dictionary while the odd ones
choose the names.
On LingBot-VLA the $74$ concepts come from each arm's gripper, its contact with the pot or the bell,
and the measured motion of its end effector.
The fitting episodes of Section~\ref{sec:setup} are split frame by frame: even frames fit the
dictionary and odd frames become the naming frames, so that no frame does both jobs.
On LingBot-VLA the split is by episode instead, with even episodes fitting the dictionary and odd
ones naming its basis features.
The autoencoder sees every token of a frame except two, which the policy attends to in every image
and whose vectors are far longer than the rest.
A decision concept is taken at the action position that writes its axis.
A state concept is taken at the last token of the instruction and at the first action position,
where the policy keeps it.
Two dictionaries are fitted per suite from two random starts, and a rule fixed before the fit picks
the one reported: the sparsity weight whose average number of basis features on per vector is nearest
fifty.

The clamp measures the share of steps whose action lies on the concept's side, and how far the arm
moves along that axis in twenty steps.
For each decision concept the clamped basis feature is the one whose name holds with the highest recall.
Where no name of that concept holds, the clamp goes to the basis feature that agrees best with the
concept on the naming frames, and only while its precision stays above one half.
Other live basis features clamped to the same multiple make the first control, a random unit direction of
equal norm the second, and a direction fitted by supervised regression the third
\citep{alain2016probes, turner2023steering}.
No name enters those three.
The fourth control is the summed direction of the named basis features of the concept.

Fitting the alarm takes four folds over the twelve initial states of the fitting episodes, and a
fold scores only the states that it did not see.
Out of fold we choose both the regularization weight and the alarm threshold.
The fused monitor of Table~\ref{tab:detect} is the largest of three scores fitted on the same
inputs: one linear, one per action position, and one tree.
We report the episode AUROC with a bootstrap interval and the share of successful episodes with a
false alarm.
The delay of an alarm is the number of steps from the simulator's failure onset to that alarm.

On the physical robot a label that a measured error could flip is left unknown, and a grasp carries
its object only when the object's visible bottom rises by at least $5$\,cm.
On $30$ demonstrations no label is wrong, and no attempt's count of completed sub-goals disagrees with
the operator's.
Twelve state concepts give the arm's motion, now and over the next half second, along $\pm x$,
$\pm y$, up and down.
The decision concepts are the sign of each axis and the hand command, open, half-closed or closed, at
each step of the chunk, and the next action is also read at the last token of the instruction, the
position at which OpenVLA decides its first action.
As on OpenVLA, a concept is counted when it is read at one token, a state at the last token of the
instruction or at the first action token and a decision at the token that decides it, and concepts
true at every patch or every action token are counted apart (Table~\ref{tab:names_robot}).
The held-out attempts are those whose number leaves a remainder of three when
divided by four, five per task, and the frames of the other attempts alternate between fitting and
naming.
The monitor is scored leaving one attempt out, and within each fold the regularization weight and the
threshold are chosen out of fold on the other attempts, as above.
The failure onset of a failed attempt is the earlier of the last time the hand lost its grip on the
object of the unfinished sub-goal, and the step, counted from the start of that sub-goal, by which
nine in ten successful attempts had reached a stage that the failed one never reaches.

\section{Detailed results}
\label{app:results}
\label{app:lingbot}
This appendix gives the full results behind Section~\ref{sec:experiments}, one measure per
subsection, first for the policies in simulation and then for $\pi_{0.5}$ on the physical robot.

\subsection{Sparsity}
\label{app:res:sparsity}
On OpenVLA about fifty of the $16{,}384$ basis features are on at one vector of Spatial, and $41$ and $42$ on
Object and Goal.
On Spatial they leave $19.5\%$ of the variance unexplained on the fitting frames and $22.3\%$ at unseen
initial states.
On the action expert of LingBot-VLA $78$ to $90$ of the $3{,}072$ basis features are on, leaving $8.7$ to
$10.8\%$ of the variance unexplained on the naming frames.
On $\pi_{0.5}$ at layer $11$, $66$ of the $8{,}192$ backbone basis features and $125$ of the $4{,}096$
action-expert basis features are on, leaving $24.6\%$ and $19.3\%$ of the variance unexplained on the fitting
layouts and $26.3\%$ and $24.6\%$ on held-out layouts.
Two random starts share $81$ to $83\%$ of the named basis features of OpenVLA, $73\%$ of them under the same
name, and agree on the unexplained variance within $0.2$ points on LingBot-VLA and to three decimals on
$\pi_{0.5}$.

On the physical robot, with the mean removed, $39$ and $30$ of the $8{,}192$ backbone basis features are on
at one vector of the stove and cup tasks, leaving $2.1\%$ and $1.5\%$ of the variance unexplained.
The action expert keeps $45$ and $57$ basis features on, leaving $43\%$ and $31\%$ of the variance within a
denoising step unexplained.
No basis feature goes unused, and a second random start recovers $86$ to $100\%$ of the basis features whose name
holds, $84$ to $91\%$ of them under the same name.

\subsection{Naming}
\label{app:res:names}
\begin{figure}[ht]
\centering
\includegraphics[width=\textwidth]{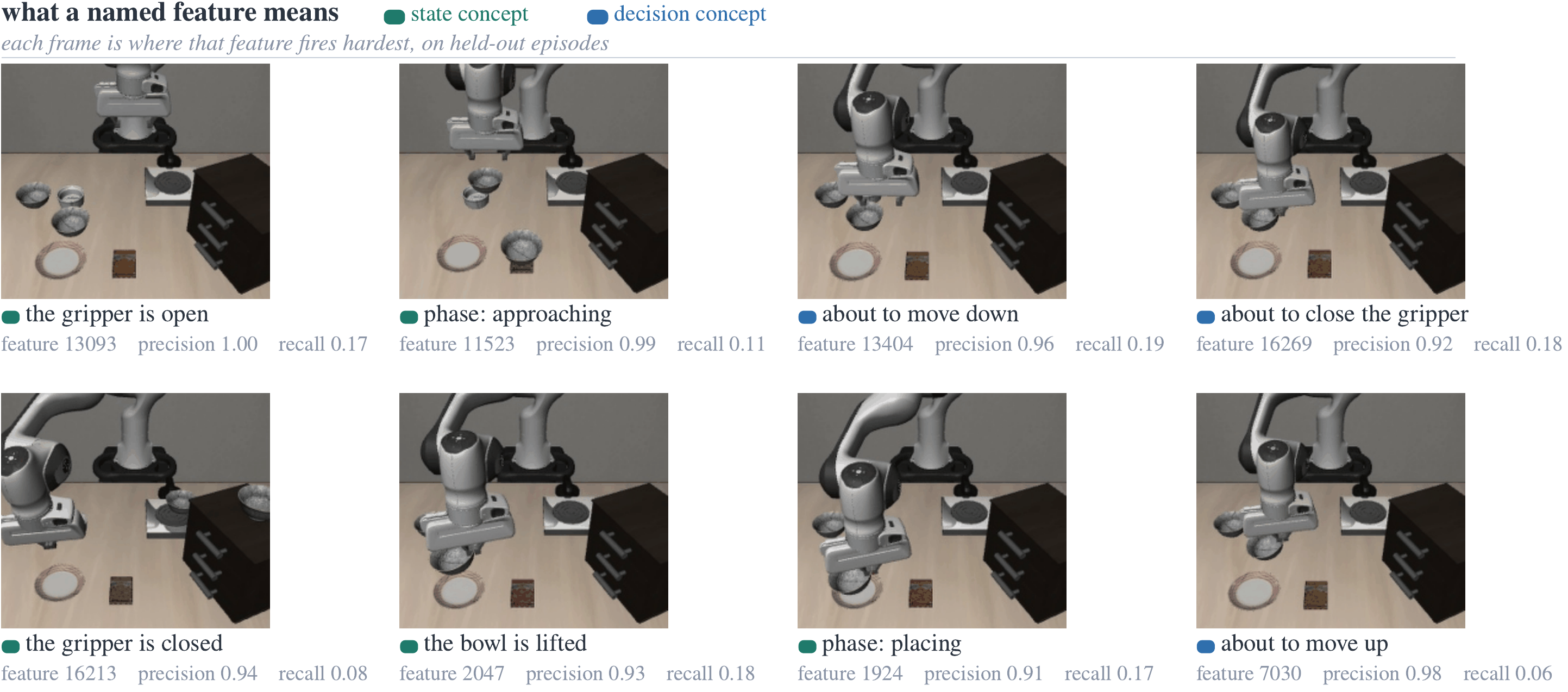}
\caption{Eight concepts that a basis feature of the LIBERO-Spatial dictionary earns a name for, each
beside the frame at which that basis feature fires hardest over the recorded held-out episodes.
Precision and recall are the ones the naming rule of Section~\ref{sec:method:name} measures on
episodes that took no part in choosing the name.}
\label{fig:concepts}
\end{figure}

\begin{figure}[ht]
\centering
\includegraphics[width=0.85\textwidth]{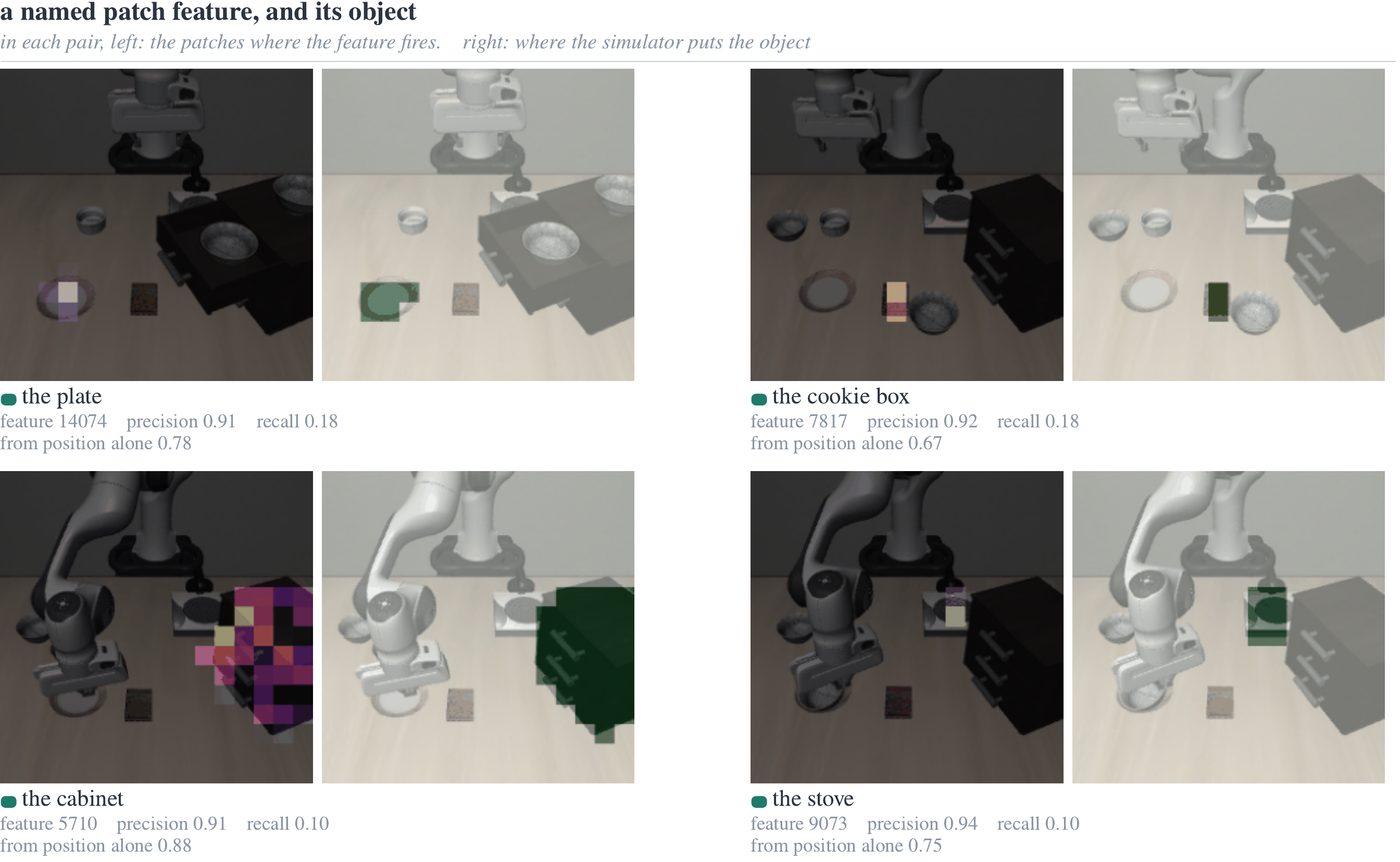}
\caption{Four basis features of the dictionary at layer twenty, each named for the object in an
image patch. In each pair, left is the set of patches on which the basis feature fires on a held-out
frame, and right is where the simulator puts that object.}
\label{fig:patches}
\end{figure}

\textbf{OpenVLA:}
Figure~\ref{fig:concepts} shows eight named concepts at the frames where their basis features fire hardest,
and Figure~\ref{fig:patches} four basis features named for the object in an image patch, each more precise
than the position of its patch alone.
Over a held-out attempt the named basis features follow the phases of its task with a median precision per
concept of $0.88$ to $1.00$.

\textbf{LingBot-VLA:}
Table~\ref{tab:names_lingbot} counts the concepts whose name holds on held-out attempts, for both
random starts of the dictionary and for every control.
A concept counts when at least one unit of the row passes the rule, and raw coordinates are counted
for each sign, since a coordinate can follow a concept when it rises or when it falls.
The dictionary names $37$ to $39$ of the $74$ concepts of lift\_pot and $16$ to $23$ of click\_bell,
at least as many as any control.
Its lead is widest on the state of the robot, with $9$ to $10$ state concepts on lift\_pot against $4$,
while raw coordinates come closer on the coming motion, which many directions of an action expert
already follow.
The named basis features follow held-out attempts with a median precision per concept of $0.74$ and $0.75$.

\begin{table}[H]
\centering
\small
\caption{Concepts of LingBot-VLA named on held-out attempts, at layer $22$ of its action expert.
Each task has $74$ concepts, taken for each arm. The largest count of each column is in bold.}
\label{tab:names_lingbot}
\begin{tabular}{@{}lcccc@{}}
\toprule
& \multicolumn{2}{c}{lift\_pot} & \multicolumn{2}{c}{click\_bell} \\
\cmidrule(lr){2-3}\cmidrule(l){4-5}
Units & Decision & State & Decision & State \\
\midrule
Dictionary, start 1 (ours) & 28 & 9  & 11 & \textbf{5} \\
Dictionary, start 2 (ours) & \textbf{29} & \textbf{10} & \textbf{19} & 4 \\
Raw coordinates, positive  & 25 & 4 & 9  & 3 \\
Raw coordinates, negative  & 18 & 4 & 11 & 3 \\
Random directions          & 10 & 4 & 10 & 2 \\
Shuffled labels            & 0  & 0 & 0  & 0 \\
\bottomrule
\end{tabular}
\end{table}

\textbf{$\pi_{0.5}$ in LIBERO:}
Table~\ref{tab:names_pi} counts the distinct concepts whose name holds on held-out layouts, over the
dictionaries of both streams at layers $7$, $11$ and $15$, and its last row counts the concepts that
neither raw coordinates nor random directions name.
The dictionary names $42$ to $53$ concepts per task, $15$ to $17$ decisions and $26$ to $36$ states,
against $13$ to $16$ for raw coordinates and $8$ to $11$ for random directions, and $28$ to $37$ of them
are named by the dictionary alone.
Every one of the $70$ decision concepts taken per step of the chunk has a basis feature whose precision holds
on held-out layouts, and a linear probe reads each of them at an AUROC of $0.994$ to $1.000$ from layer
$7$ on.
Basis features named for an object again beat the position of their patch, the basket at a precision of
$0.92$ against $0.09$ and the arm at $0.92$ against $0.04$.
The named basis features of layer $11$ follow held-out attempts with a median precision per concept of
$0.62$ to $0.85$.

\begin{table}[H]
\centering
\small
\caption{Concepts of $\pi_{0.5}$ named on held-out layouts, over both streams and the layers $7$,
$11$ and $15$.}
\label{tab:names_pi}
\begin{tabular}{@{}lcccc@{}}
\toprule
Units & LIBERO-10 & Goal & Object & Spatial \\
\midrule
Dictionary (ours)             & \textbf{53} & \textbf{48} & \textbf{42} & \textbf{50} \\
\quad of which decisions       & 17 & 16 & 16 & 15 \\
\quad of which states          & 36 & 32 & 26 & 35 \\
Raw coordinates               & 16 & 14 & 16 & 13 \\
Random directions             & 11 & 8 & 8 & 8 \\
\midrule
Named by the dictionary alone & 36 & 32 & 28 & 37 \\
\bottomrule
\end{tabular}
\end{table}

\begin{figure}[ht]
\centering
\includegraphics[width=\textwidth]{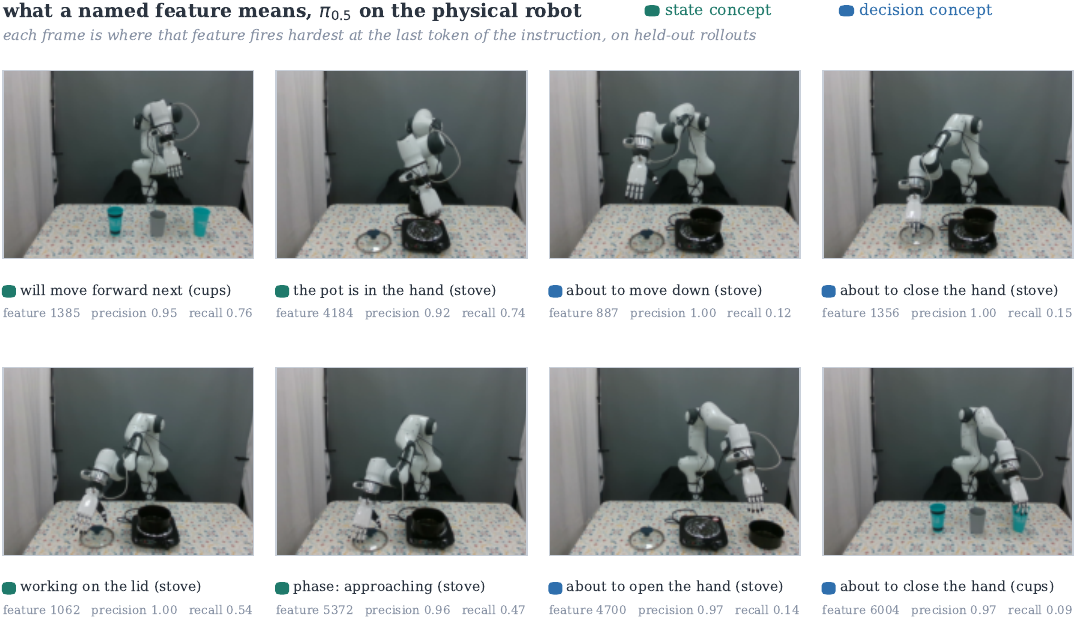}
\caption{Eight concepts that a basis feature of $\pi_{0.5}$ on the physical robot earns a name for, read at the
last token of the instruction in its backbone at layer $11$, each beside the camera frame at which that
basis feature fires hardest over the held-out attempts of its task. The four decisions about the next action are
all that the dictionary names there; the four states are those named with the highest recall, one per
concept. Precision and recall are measured on attempts that took no part in choosing the name.}
\label{fig:concepts_robot}
\end{figure}

\textbf{$\pi_{0.5}$ on the physical robot:}
At the last token of the instruction, $24$ basis features keep their names on held-out stove attempts for
$11$ state concepts, and $4$ for $3$ decisions about the next action: closing the hand, opening it and
moving down.
On the cups $11$ basis features keep them for $5$ state concepts, and $1$ for closing the hand.
Raw coordinates, random directions and shuffled labels name none of these concepts, so the dictionary
wins all $14$ of the stove task and all $6$ of the cups, and Figure~\ref{fig:concepts_robot} shows eight
of them.
A linear probe confirms that every concept is present.
On held-out attempts it reads the concepts at a median AUROC of $0.998$ and $0.992$ in the backbone of
the two tasks, and of $0.997$ to $0.999$ and $0.993$ to $0.998$ in the action expert, and it reads the
next action at the last token of the instruction, before the action expert runs, at $0.97$ to $1.00$.
In a scan over layers $7$, $11$ and $15$ the probe finds the concepts at every layer, at median AUROCs of
$0.988$ to $0.999$ in both streams and both tasks.

The action expert is harder to name.
At its last denoising step its input is already close to the finished chunk, and $57$ to $59\%$ of its
variance copies the action that it writes, so a name there mixes the decision with its fed-back input,
as the memory concepts of OpenVLA do.
There the dictionary names $5$ and $8$ concepts against $1$ and $2$ for raw coordinates, and at the
first denoising step, whose input is noise, it names none while raw coordinates name $4$ on each task.
Concepts true at every action token, such as the open hand, are counted apart, since the first
principal component of the action expert holds $34\%$ of its variance and reads the hand command at an
AUROC of $1.00$.

Table~\ref{tab:names_robot} counts the units of $\pi_{0.5}$ on the physical robot whose name holds on
held-out attempts, with the concepts they cover, at layer $11$, on the labels with the carry rule of
Appendix~\ref{app:setup}.
As OpenVLA's counts, the first four columns hold concepts read at one token: in the backbone, a state
or the next action at the last token of the instruction; in the action expert, a state at the first
action token and a decision at the token that decides it.
The last two columns count apart the concepts true at every patch or every action token, which raw
coordinates and random directions pass by the hundreds in the action expert.
Raw coordinates and random directions are counted for both signs.
The last row of each task counts the concepts of one token won by the dictionary against those won by
raw coordinates: a kind of unit wins a concept when only it holds a name for it, or when both do and
its best held-out recall is higher.

\begin{table}[H]
\centering
\small
\caption{Units of $\pi_{0.5}$ on the physical robot named on held-out attempts, with the number of
concepts they cover in brackets, at layer $11$. B: the backbone at the last token of the
instruction; AE: the action expert at its last and first denoising steps; coord.: raw coordinates. The last two columns count
the concepts true at every patch (backbone) or every action token (action expert, last step).}
\label{tab:names_robot}
\setlength{\tabcolsep}{3.5pt}
\begin{tabular}{@{}llcccccc@{}}
\toprule
& & \multicolumn{4}{c}{concepts of one token} & \multicolumn{2}{c}{every patch or token} \\
\cmidrule(lr){3-6}\cmidrule(l){7-8}
Task & Units & B, state & B, next action & AE, last & AE, first & B & AE, last \\
\midrule
\multirow{5}{*}{stove}
& Dictionary (ours) & \textbf{24} (\textbf{11}) & \textbf{4} (\textbf{3}) & \textbf{5} (\textbf{5}) & 0 & 122 (11) & 61 (22) \\
& Raw coordinates   & 0 & 0 & 1 (1) & \textbf{7} (\textbf{4}) & 2 (2) & 942 (23) \\
& Random directions & 0 & 0 & 0 & 0 & 78 (4) & 3{,}231 (21) \\
& Shuffled labels   & 0 & 0 & 0 & 0 & 0 & 0 \\
& Dictionary vs coord. & \multicolumn{2}{c}{14 vs 0} & 5 vs 1 & 0 vs 4 & & \\
\midrule
\multirow{5}{*}{cups}
& Dictionary (ours) & \textbf{11} (\textbf{5}) & \textbf{1} (\textbf{1}) & \textbf{14} (\textbf{8}) & 0 & 216 (14) & 127 (17) \\
& Raw coordinates   & 0 & 0 & 2 (2) & \textbf{4} (\textbf{4}) & 8 (3) & 1{,}144 (22) \\
& Random directions & 0 & 0 & 0 & 0 & 174 (5) & 3{,}864 (19) \\
& Shuffled labels   & 0 & 0 & 0 & 0 & 0 & 0 \\
& Dictionary vs coord. & \multicolumn{2}{c}{6 vs 0} & 6 vs 2 & 0 vs 4 & & \\
\bottomrule
\end{tabular}
\end{table}

\subsection{Steering}
\label{app:res:steer}
\begin{figure}[ht]
\centering
\includegraphics[width=\textwidth]{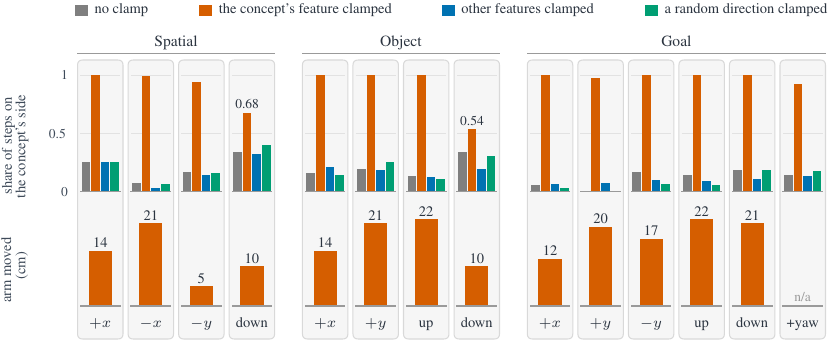}
\caption{Clamping one basis feature of OpenVLA in closed-loop rollouts on the unseen task, from six to
eight held-out initial states per suite. Each box is one clamped concept. Its upper bars give the
share of steps whose action lies on the concept's side, with no clamp, with the concept's basis feature
clamped, and under the two controls. Its lower bar gives how far the clamped arm ends from the
unclamped one along the concept's axis after twenty steps, and the turn has no such distance.
Table~\ref{tab:steer} lists every value.}
\label{fig:steer}
\end{figure}

\begin{figure}[t]
\centering
\includegraphics[width=\textwidth]{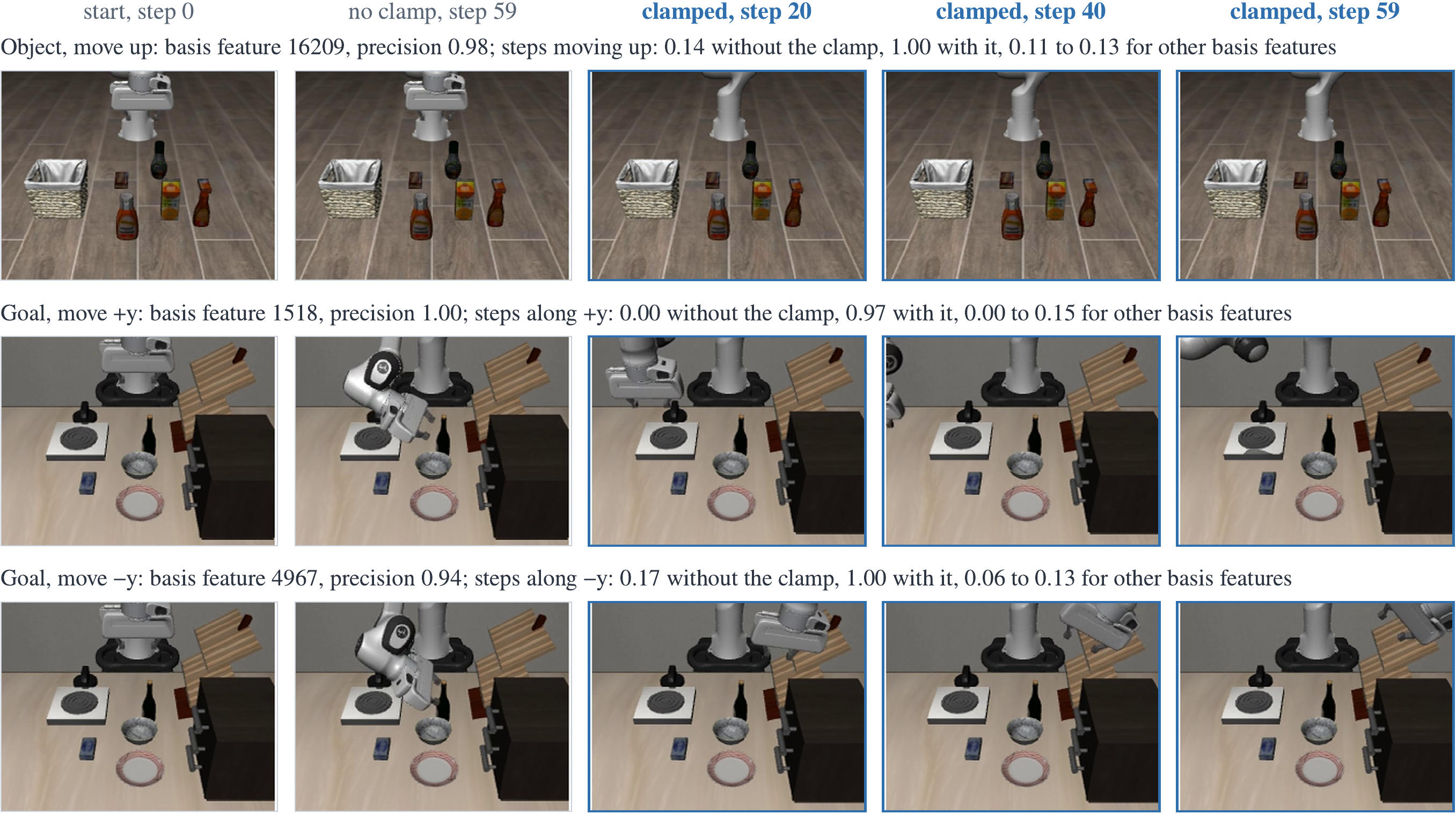}
\caption{Clamping one named basis feature of OpenVLA in closed loop, on the held-out task from its
first initial state. Each row starts from the same frame; without the clamp the policy follows its own
course, and with the clamp (blue frames) the arm moves the way the concept says. Shares of steps are
averaged over six to eight initial states (Table~\ref{tab:steer}).}
\label{fig:steer_frames}
\end{figure}

Table~\ref{tab:steer} lists every clamped rollout of Figure~\ref{fig:steer}.
Figure~\ref{fig:steer_frames} shows three of these rollouts frame by frame.
For twelve of the fourteen motion concepts, clamping one basis feature of OpenVLA raises the share of steps on
the concept's side from $0.00$--$0.25$ to $0.92$--$1.00$ and moves the arm $5$ to $22$\,cm along the
concept's axis within twenty steps.
On these twelve, one basis feature of a dictionary fitted without labels steers about as strongly as a
direction fitted to the concept's labels by regression, which reaches $0.95$ to $1.00$.
Other live basis features clamped to the same multiple, and a random direction of equal norm, never raise the
share by more than $0.08$.
For the downward move on Spatial and Object one basis feature raises the share from $0.34$ and $0.35$ to
$0.68$ and $0.54$, and the summed direction of the concept's named basis features raises it to $1.00$.

\begin{table}[h]
\centering
\small
\caption{The values drawn in Figure~\ref{fig:steer}. Closed-loop rollouts on the unseen task, from
six to eight held-out initial states. Each number is the share of steps whose action lies on the
named concept's side. The summed direction adds the named basis features of the concept, and the fitted
direction is fitted to the concept's labels by regression; a dash marks a concept for which no
summed direction was formed. The last column is how far the arm clamped on one basis feature ends from the unclamped
one along the concept's axis after twenty steps.}
\label{tab:steer}
\setlength{\tabcolsep}{4pt}
\begin{tabular}{@{}llcccccccc@{}}
\toprule
& & & & \multicolumn{2}{c}{null controls} & \multicolumn{2}{c}{directions} & \\
\cmidrule(lr){5-6}\cmidrule(lr){7-8}
suite & concept & no clamp & \begin{tabular}[b]{@{}c@{}}one basis\\feature\end{tabular} & \begin{tabular}[b]{@{}c@{}}other basis\\features\end{tabular} & random & summed & fitted & arm moves \\
\midrule
Spatial & move $+x$   & 0.25 & \textbf{1.00} & 0.25 & 0.25 & 1.00 & 1.00 & 14\,cm \\
Spatial & move $-x$   & 0.08 & \textbf{0.99} & 0.03 & 0.07 & 0.20 & 0.97 & 21\,cm \\
Spatial & move $-y$   & 0.17 & \textbf{0.94} & 0.14 & 0.16 & --   & 1.00 & 5\,cm \\
Spatial & move down   & 0.34 & 0.68 & 0.33 & 0.40 & 1.00 & 0.97 & 10\,cm \\
Object  & move $+x$   & 0.16 & \textbf{1.00} & 0.21 & 0.14 & 0.96 & 1.00 & 14\,cm \\
Object  & move $+y$   & 0.19 & \textbf{1.00} & 0.19 & 0.26 & 1.00 & 1.00 & 21\,cm \\
Object  & move up     & 0.14 & \textbf{1.00} & 0.13 & 0.11 & 1.00 & 1.00 & 22\,cm \\
Object  & move down   & 0.35 & 0.54 & 0.20 & 0.31 & 1.00 & 1.00 & 10\,cm \\
Goal    & move $+x$   & 0.06 & \textbf{1.00} & 0.06 & 0.03 & 0.97 & 0.95 & 12\,cm \\
Goal    & move $+y$   & 0.00 & \textbf{0.97} & 0.08 & 0.01 & 1.00 & 1.00 & 20\,cm \\
Goal    & move $-y$   & 0.17 & \textbf{1.00} & 0.10 & 0.06 & 0.97 & 0.99 & 17\,cm \\
Goal    & move up     & 0.14 & \textbf{1.00} & 0.10 & 0.06 & 1.00 & 1.00 & 22\,cm \\
Goal    & move down   & 0.19 & \textbf{1.00} & 0.11 & 0.19 & 1.00 & 1.00 & 21\,cm \\
Goal    & turn $+$yaw & 0.14 & \textbf{0.92} & 0.14 & 0.18 & 0.98 & 0.99 & -- \\
\bottomrule
\end{tabular}
\end{table}

\begin{figure}[t]
\centering
\includegraphics[width=\textwidth]{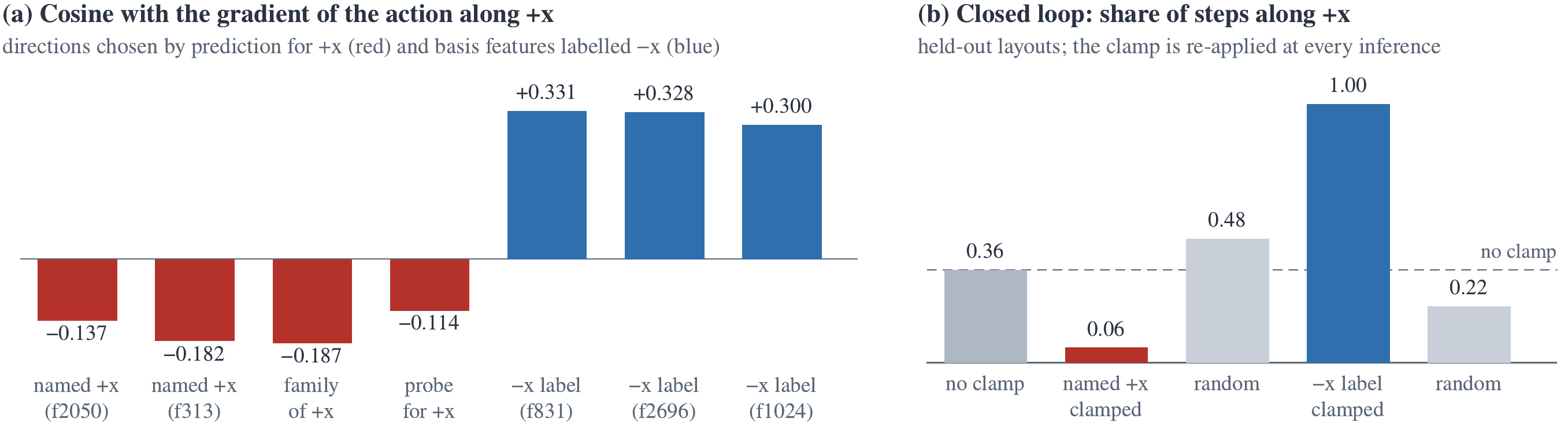}
\caption{The sign inversion of $\pi_{0.5}$ in LIBERO, at layer $11$ of the action expert. (a) Every
direction chosen by prediction for $+x$ points against the gradient of the action, while basis features
labelled $-x$ below the precision floor point along it. (b) Clamping the basis feature named for $+x$
lowers the share of steps along $+x$, and clamping one labelled $-x$ raises it to $1.00$; each random
bar is the control of the clamp to its left.}
\label{fig:pi_inversion}
\end{figure}

On $\pi_{0.5}$ in LIBERO the name of a decision basis feature gives the axis of the action, and its
clamp gives the opposite sign (Figure~\ref{fig:pi_inversion}).
At layer $11$ of the action expert, the direction that causes a move is the gradient of the action with
respect to the activation, taken through the flow sampler.
Every direction chosen by prediction for the same decision is anti-aligned with it: all nine named basis
features tested, at precisions of $0.906$ to $0.990$, the summed directions of their four families and
the four supervised probes, for $+x$ on two suites, for moving down and for closing the gripper, at
cosines of $-0.08$ to $-0.26$.
The inversion therefore belongs to how $\pi_{0.5}$ encodes its decisions, since a probe trained with
labels shows it as well as the dictionary does.
For $+x$, $14$ of the $20$ basis features most aligned with the gradient carry the label $-x$, but at a
precision of $0.04$ to $0.37$, below the floor of the naming rule.
Clamping one of them in closed loop on held-out layouts raises the share of steps along $+x$ from
$0.36$ to $1.00$, against $0.22$ for random basis features, and drops task success from $1.00$ to $0.00$.
Clamping the basis feature named for $+x$ instead lowers that share to $0.06$, against $0.48$ for random
basis features.
Open loop, three of these basis features move the action along $+x$ at slopes of $0.0063$, $0.0064$ and
$0.0061$ per unit of clamp norm, linear in its size, and steering along the gradient itself raises the share
of steps along $+x$ from $38\%$ to $88\%$ at a fifth of the norm of the activation.
The same basis features in the same order lead a second suite.

\subsection{Failure detection}
\label{app:res:detect}
\begin{table}[ht]
\centering
\small
\caption{Failure detection in full, with the intervals behind Table~\ref{tab:detect}. AUROC is per episode, with its $95\%$ interval; false alarms are the
share of successful episodes with an alarm. Dense activation: the same monitor on the activation itself.
Ours reads the named basis features; fused takes the largest of their z-scored linear, per-position and tree
scores. LingBot-VLA is scored on its opening frames, $\pi_{0.5}$ in LIBERO over both streams at layers $7$, $11$ and
$15$, and the robot at layer $11$ of its backbone, leaving one attempt out ($8$ successes).
$^\dagger$Best robot monitor fixed before any fit (Appendix~\ref{app:res:detect}). Best AUROC per task
in bold.}
\label{tab:detect_full}
\setlength{\tabcolsep}{5pt}
\begin{tabular}{@{}c l l c c@{}}
\toprule
& Task & Monitor & AUROC $\uparrow$ & False alarms $\downarrow$ \\
\midrule
\multirow{9}{*}{\rotatebox[origin=c]{90}{OpenVLA}}
& \multirow{3}{*}{Spatial}
  & Dense activation     & \textbf{1.00} {\scriptsize[1.00, 1.00]} & 2\% \\
& & Ours                 & \textbf{1.00} {\scriptsize[0.99, 1.00]} & 3\% \\
& & Ours, fused          & \textbf{1.00} {\scriptsize[0.99, 1.00]} & 2\% \\
\cmidrule(l){2-5}
& \multirow{3}{*}{Object}
  & Dense activation     & 0.91 {\scriptsize[0.81, 0.98]} & 7\% \\
& & Ours                 & 0.89 {\scriptsize[0.78, 0.96]} & 5\% \\
& & Ours, fused          & \textbf{0.93} {\scriptsize[0.87, 0.98]} & 9\% \\
\cmidrule(l){2-5}
& \multirow{3}{*}{Goal}
  & Dense activation     & 0.92 {\scriptsize[0.84, 0.99]} & 7\% \\
& & Ours                 & 0.92 {\scriptsize[0.83, 0.99]} & 5\% \\
& & Ours, fused          & \textbf{0.97} {\scriptsize[0.93, 1.00]} & 2\% \\
\midrule
\multirow{4}{*}{\rotatebox[origin=c]{90}{\shortstack{LingBot-\\VLA}}}
& \multirow{2}{*}{lift\_pot}
  & Dense activation     & \textbf{0.60} {\scriptsize[0.52, 0.67]} & 4\% \\
& & Ours                 & 0.58 {\scriptsize[0.50, 0.65]} & 4\% \\
\cmidrule(l){2-5}
& \multirow{2}{*}{click\_bell}
  & Dense activation     & 0.63 {\scriptsize[0.48, 0.77]} & 3\% \\
& & Ours                 & \textbf{0.72} {\scriptsize[0.63, 0.81]} & 2\% \\
\midrule
\multirow{2}{*}{\rotatebox[origin=c]{90}{$\pi_{0.5}$}}
& \multirow{2}{*}{4 tasks}
  & Dense activation     & 0.988 {\scriptsize[0.973, 0.999]} & 6\% \\
& & Ours                 & \textbf{0.991} {\scriptsize[0.980, 1.000]} & 3\% \\
\midrule
\multirow{8}{*}{\rotatebox[origin=c]{90}{\shortstack{$\pi_{0.5}$,\\robot}}}
& \multirow{4}{*}{stove}
  & Dense activation     & 0.74 {\scriptsize[0.45, 0.93]} & 13\% \\
& & Ours                 & 0.70 {\scriptsize[0.44, 0.95]} & 13\% \\
& & Ours, fused          & 0.68 {\scriptsize[0.39, 0.93]} & 38\% \\
& & Ours, averaged$^\dagger$ & \textbf{0.84} {\scriptsize[0.64, 1.00]} & 13\% \\
\cmidrule(l){2-5}
& \multirow{4}{*}{cups}
  & Dense activation     & 0.61 {\scriptsize[0.32, 0.85]} & 13\% \\
& & Ours                 & 0.60 {\scriptsize[0.33, 0.86]} & 0\% \\
& & Ours, fused          & 0.43 {\scriptsize[0.20, 0.66]} & 0\% \\
& & Ours, per group$^\dagger$ & \textbf{0.70} {\scriptsize[0.49, 0.90]} & 13\% \\
\bottomrule
\end{tabular}
\end{table}

On Spatial every monitor of Table~\ref{tab:detect_full} separates failed from successful attempts at an
AUROC of $1.00$.
On Object and Goal the named basis features reach $0.89$ and $0.92$ against $0.91$ and $0.92$ for the dense
activation, with overlapping intervals over twenty to twenty-five failed episodes and $5\%$ false alarms
against $7\%$, and their fused variant scores highest, at $0.93$ and $0.97$.
The dense monitor alarms sooner on Spatial and Object, but $29$ to $38\%$ of its alarms precede the
simulator's failure onset, against $7$ to $21\%$ for the fused monitor on the named basis features.
On $\pi_{0.5}$, over $77$ held-out failures and $88$ successes, the monitor reads $2{,}503$ named basis
features of the action expert and $1{,}922$ of the backbone, for $31$ concepts.
It reaches $0.991$ [$0.980$, $1.000$] against $0.988$ [$0.973$, $0.999$] for the dense activation, a tie
within the intervals, and at its operating point it catches $99\%$ of the failures against $95\%$, with
$3\%$ false alarms against $6\%$ and a median of $10$ steps after the failure onset against $25$.
The two streams complement each other.
The action expert alone reaches $0.985$ and catches $84\%$ of the failures, and the backbone alone,
without dwells or switching rates, reaches $0.976$ at $1\%$ false alarms.
On LingBot-VLA, scored on the opening frames, the named basis features beat the dense activation on
click\_bell, $0.72$ against $0.63$, where every failed arm stops short of the bell, and tie it on
lift\_pot, $0.58$ against $0.60$, where $58$ of the $91$ failures come after the grasp, beyond the
scored frames.

On LingBot-VLA the monitors read the first eight frames of lift\_pot and six of click\_bell, where the
length of an attempt alone scores $0.50$, and on $\pi_{0.5}$ in LIBERO the named basis features are scored by
the tree monitor.

On the physical robot each fit learns from at most $11$ failures of the stove task and $14$ of the cups.
Two monitors in Table~\ref{tab:detect_full} carry a dagger: they are the best of the monitors fixed before
any fit.
The averaged monitor takes the mean of the z-scored linear, per-position and tree monitors on the named
basis features, and the per-group monitor fits one monitor per token group, image, instruction and state.
On the stove task the best monitor fixed before any fit averages the concept monitors of the backbone,
at $0.84$ [$0.64$, $1.00$], and catches $8$ of $11$ failures with one false alarm in $8$ successes, a
median of two seconds before the operator stopped the attempt.
On the cups the best reads the named basis features per token group, at $0.70$ [$0.49$, $0.90$], and catches
$8$ of $14$ failures with one false alarm, including all six attempts that failed on the second cup and
none of the four that failed on the first, which stop before any state that the monitor learned.
Under the main protocol the named basis features of the backbone reach $0.70$ and $0.60$ and the dense
activation $0.74$ and $0.61$, all with intervals that include chance, and those of the action expert
$0.61$ and $0.29$.

\textbf{Data size.}
We tested whether the number of robot attempts limits the monitor, with the design written down
before any run.
First, we draw random subsets of the LIBERO episodes of OpenVLA with the robot's counts, $11$ failed
and $8$ successful episodes or $14$ and $8$, and score each subset as the robot is scored, leaving one
episode out.
The dictionary and the names stay those of the full data, and only the monitor's training episodes
shrink.
At that size the named basis features of OpenVLA still reach a median AUROC of $0.86$ to $1.00$, against
$0.70$ and $0.60$ on the robot.
Second, we refit the robot monitors on $6$, $10$ or $14$ randomly drawn attempts.
Their AUROC does not rise with more attempts, and draws differ by about $\pm 0.1$.
The difference lies in how much of a failure is recorded.
The failed attempts of OpenVLA run to the step limit, and $64$ to $66\%$ of their frames come after the
failure onset.
The operator stops a failing robot attempt early, and only $9$ to $14\%$ of its frames, about two
seconds, come after the onset.

\subsection{Diagnosis}
\label{app:res:diag}
\begin{figure}[ht]
\centering
\includegraphics[width=\textwidth]{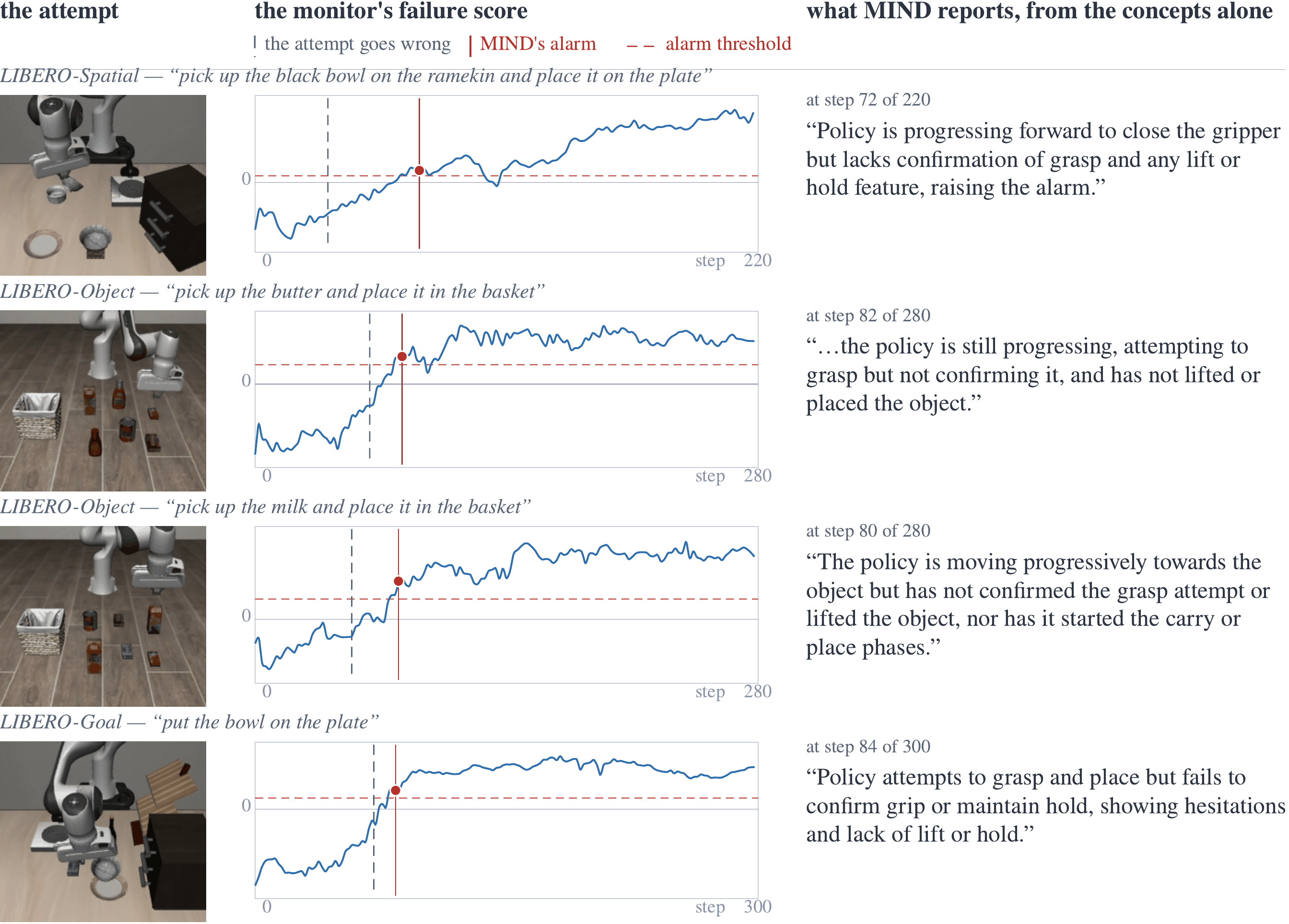}
\caption{Four failures that MIND called, one to a row: the camera frame at the step the alarm
fired, the failure score over the whole attempt, and the sentence the language model wrote from
the concept record alone. The grey line marks the step at which the simulator says the attempt
has gone wrong, and the dot marks the alarm on the score.}
\label{fig:diagnosis}
\end{figure}

On one held-out attempt at the black-bowl task of OpenVLA the alarm fires at step $66$, $32$ steps after
the failure onset.
Its evidence falls on the silence of the basis features that are on in every ordinary frame ($32\%$), the
decision to close the gripper now ($30\%$) and the forward move just written ($13\%$), while no basis feature
for the bowl grasped ever came on.
The policy decided to close the gripper before the concepts that precede that decision, as in the toy
example of Section~\ref{sec:method:cause}, and the note says so: the policy ``decided to close the
gripper five times \ldots without a lasting gripper-closed basis feature; its lifted basis feature lasted only from
step 60 to 79''.
The same concepts lead the other alarms of Spatial, and across twenty bootstrap refits of the monitor
the top concept of an alarm stays among the three largest in a median of $90$ to $95\%$ of the refits.
On the three suites $85$ to $89\%$ of the model's answers about the grasp, the hold, the placing and the
motion match the record, and the model calls $11$ of the $12$ successful attempts normal.
Each note is checked against the record, and $69$ to $88\%$ of the notes come from the model, the rest
from fixed sentences built from the same record.
The final prompt shows the model the category that a mechanical rule reads from the record, and the
model keeps that category for $14$ of $19$, $14$ of $14$ and $13$ of $15$ failures on Spatial, Object and
Goal.
Against the simulator its category is right for $2$ of $19$, $9$ of $14$ and $4$ of $15$ failures, and
the rule's for $4$, $9$ and $3$, so on OpenVLA as on $\pi_{0.5}$ the category is bounded by the
concepts that the record holds.
On $\pi_{0.5}$ in LIBERO, at layer $11$, $75\%$ of the answers match the record, but the category agrees
with the simulator for only $33\%$ of the detected failures, as often as a mechanical reading of the
record ($36\%$), since the phases that separate the failure types are not linearly present in its action
expert.

On the physical robot the labels can score $8$ of the $11$ stove failures, since the hand hides the lid
in the other three.
With the record reduced to the best basis feature of each concept, the model names the right failure for $3$
of the $8$ and a mechanical rule on the same record for $4$, against $1$ and $3$ with the full record,
and the model calls $7$ of the $8$ successes normal.
Its main confusion, a lid never grasped against a lid not placed, comes from basis features for holding the
lid that fire briefly in attempts that never grasped it.
On the cups it names the right failure for $4$ of $10$, but calls only $1$ or $2$ of the $8$ successes
normal.

\subsection{Ablations}
\label{app:res:ablation}
\begin{figure}[h]
\centering
\includegraphics[width=\textwidth]{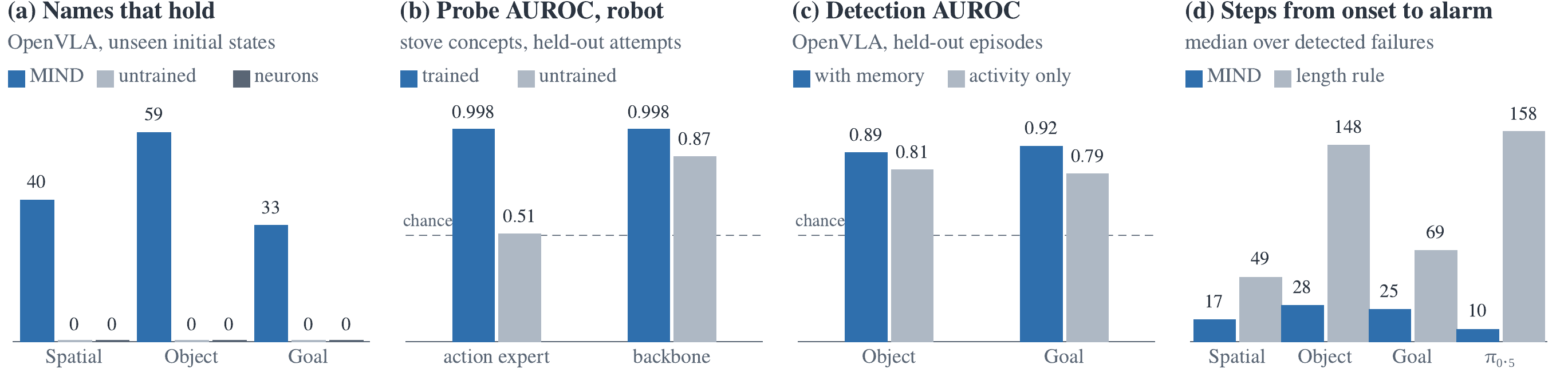}
\caption{The ablations of Table~\ref{tab:ablation}. (a) Names that hold at unseen initial states,
decision and state together. (b) Median probe AUROC over the stove concepts on the physical robot. (c)
Detection AUROC with and without the dwells and the switching rate. (d) Median steps from the failure
onset to the alarm, where lower is earlier.}
\label{fig:ablation}
\end{figure}

Table~\ref{tab:ablation} summarizes the four ablations of Section~\ref{sec:results:ablation},
Figure~\ref{fig:ablation} plots them, and this section gives their full numbers.
Each ablation replaces one part of MIND and keeps every other part, including the constants of
Appendices~\ref{app:monitor} and~\ref{app:name}.

\begin{table}[h]
\centering
\caption{Ablations. Each row replaces one part of MIND and keeps the rest unchanged. Names are
those that hold at unseen initial states, as decision / state basis features; delays are the median
steps from the failure onset to the alarm.}
\label{tab:ablation}
\small
\setlength{\tabcolsep}{4pt}
\begin{tabular}{@{}lllcc@{}}
\toprule
Ablation & Policy & Measure & MIND & Ablated \\
\midrule
Untrained policy & OpenVLA & names that hold & 15--31 / 13--28 & 0 / 0 \\
 & $\pi_{0.5}$, robot & probe AUROC, action expert & 0.997--0.999 & 0.51 \\
Neurons for the dictionary & OpenVLA & names that hold & 15--31 / 13--28 & 0 / 0 \\
 & $\pi_{0.5}$, LIBERO & one-step decisions named & 6 & 1 \\
Record without memory & OpenVLA & AUROC, Object / Goal & 0.89 / 0.92 & 0.81 / 0.79 \\
Length rule for the monitor & OpenVLA, $\pi_{0.5}$ & delay after onset & 10--28 & 49--158 \\
\bottomrule
\end{tabular}
\end{table}

\textbf{Untrained policy:}
The dictionary is fitted on the activations of a copy of OpenVLA whose weights were never trained.
It leaves $84\%$ of its basis features unused, and none of the rest earns a name that holds.
On the robot a linear probe on an untrained copy of $\pi_{0.5}$ finds the stove concepts at a median
AUROC of $0.51$ in the action expert at both denoising steps, the level of shuffled labels, against
$0.997$ to $0.999$ when trained.
In the backbone it finds them at $0.87$ against $0.998$, since the untrained backbone still receives the
image and the joint readings written into the instruction.

\textbf{Neurons for the dictionary:}
The naming rule is applied to the neurons of the feed-forward layers in place of the basis features.
On OpenVLA the neurons name no concept, and for $29$ of $30$ pairs of concept and layer the best basis
feature beats the best neuron, typically by a factor of three in precision.
At the same layers of the action expert of $\pi_{0.5}$ in LIBERO the neurons name $6$ to $8$ concepts per
task against $7$ to $10$ for the dictionary, but all except one of their concepts are true at every
action token.
On decisions tied to one step of the chunk the neurons name $1$ and the dictionary $6$.

\textbf{Record without memory:}
The monitor keeps only the activity of each named basis feature and drops its dwells and its switching
rate.
Its AUROC then falls from $0.89$ to $0.81$ on Object and from $0.92$ to $0.79$ on Goal.

\textbf{Length rule for the monitor:}
A failed attempt runs longer than a successful one, but the step index is never an input of the
monitor.
A rule that alarms once an attempt outlasts $95\%$ of the successful ones fires a median of $49$, $148$
and $69$ steps after the failure onset on Spatial, Object and Goal, where the named basis features alarm
after $17$, $28$ and $25$.
On $\pi_{0.5}$ in LIBERO the rule fires after $158$ steps and the named basis features of both streams
after $10$.

\end{document}